%% file: camera-ready.tex
\documentclass[letterpaper]{article} 
\usepackage{aaai2026}  
\usepackage{times}  
\usepackage{helvet}  
\usepackage{courier}  
\usepackage[hyphens]{url}  
\usepackage{graphicx} 
\usepackage{natbib}  
\usepackage{caption} 
\usepackage{algorithm}
\usepackage{algorithmic}

\usepackage{newfloat}
\usepackage{listings}
\DeclareCaptionStyle{ruled}{labelfont=normalfont,labelsep=colon,strut=off} 
\floatstyle{ruled}
\newfloat{listing}{tb}{lst}{}
\floatname{listing}{Listing}
\title{When Human Preferences Flip: An Instance-Dependent Robust Loss for RLHF}
\author {
    Yifan Xu,\textsuperscript{\rm 1}~
    Xichen Ye,\textsuperscript{\rm 2}~
    Yifan Chen,\textsuperscript{\rm 1,}\thanks{
    Correspondence to: Yifan Chen \textlangle yifanc@hkbu.edu.hk\textrangle.}~
    Qiaosheng Zhang\textsuperscript{\rm 3}
}
\affiliations {
    \textsuperscript{\rm 1}Hong Kong Baptist University\\
    \textsuperscript{\rm 2}Fudan University\\
    \textsuperscript{\rm 3}Shanghai Artificial Intelligence Laboratory
}

\usepackage{multirow}
\usepackage{booktabs}
\usepackage{graphicx}
\usepackage{caption}
\usepackage{subcaption}
\usepackage[flushleft]{threeparttable}
\usepackage{siunitx} 

\usepackage{mathtools}
\usepackage{comment}
\usepackage{bm}
\usepackage{dsfont}
\usepackage{amsmath}
\usepackage{amssymb}
\usepackage{amsfonts}
\usepackage{amsthm}
\usepackage{pifont}
\usepackage{lipsum}
\usepackage{enumitem}

\usepackage{cleveref}

\usepackage{tikz}

\theoremstyle{plain}
\newtheorem{theorem}{Theorem}[section]
\newtheorem{proposition}[theorem]{Proposition}
\newtheorem{lemma}[theorem]{Lemma}

\theoremstyle{definition}

\newtheorem{assumption}[theorem]{Assumption}

\theoremstyle{remark}

\def\spacingset#1{\renewcommand{\baselinestretch}%
{#1}\small\normalsize}

\input{math_commands}

\begin{document}

\maketitle


\begin{abstract}
Quality of datasets plays an important role in large language model (LLM) alignment.
In collecting human feedback, however, \emph{preference flipping} is ubiquitous and causes corruption in data annotation;
the issue necessitates the alignment algorithms with improved robustness against potential flipped pairs.
To this end, this paper introduces a Flipping-Aware Direct Preference Optimization (FA-DPO) algorithm tailored to preference flipping from a reinforcement learning with human feedback (RLHF) perspective. 
We dissect the inherent human intention model and the preference flipping mechanism introduced by external factors as two distinct stages;
in the latter, we introduce an instance-dependent flipping probability on the basis of the Bradley-Terry (BT) model.
Further, by leveraging features relevant to preference annotation, we capture uncertainty in judgments and model preference flipping patterns.
In practice, we design a simple yet efficient iterative optimization algorithm compatible with the original RLHF and DPO algorithms.
In our experiments, we investigate the instance-dependent preference flipping model under multiple circumstances for evaluation of our proposed method, as well as other baseline methods.
\end{abstract}

\section{Introduction}
\label{sec: intro}
Alignment has been identified as a crucial approach contributing to the remarkable capacity of large language models (LLMs) to understand human intentions. 
In general, this approach enables LLMs to produce responses that align well with human expectations and reduce toxic generations \citep{achiam2023gpt, dubey2024llama}.
Among existing alignment paradigms, a notable one is the reinforcement learning from human feedback (RLHF), which, along with its numerous variants, has attracted significant attention~\citep{ouyang2022training, casper2023open}.

Despite the effectiveness of current RLHF methods, they implicitly suffer from noise in human feedback data \citep{zheng2023judging, gao2024impact}. 
For example, \citet{gao2024impact} reported that a $10\%$ increase in preference flipping ratios can result in a $30\%$ decrease in alignment performance, as measured by \emph{the win rate} (defined in \Cref{sec:exp-setup}).
As noise is inevitably introduced during data collection process, robust alignment algorithms bring the benefits of not only defending potential dataset attacks, but also reducing the costs of collecting clean data.
One existing genre of robust learning approaches leverages noise information inferred from existing data~\citep{xia2021sample, song2022learning}, 
and this characteristic is especially attractve in the contexts of direct alignment algorithms such as direct preference optimization (DPO; \citealt{rafailov2024direct}), where training relies on a fixed offline dataset. 

In tackling robustness challenges in RLHF, previous research focuses on a simplified scenario where preferences are randomly flipped at a \emph{fixed} rate \citep{chowdhury2024provably, wu2024towards, cheng2024rime, liang2024robust}. 
However, in practical applications, it is ungrounded to assume the possibility of annotation errors is independent of the specific content being annotated.
In this work, we instead investigate a setting of ``instance-dependence'' \citep{xia2020part, liu2023identifiability},
which poses greater challenges as the noise distribution can vary significantly across different samples.
Therefore, in this work, we aim to answer the following question:
\begin{center}
    \textit{How can we properly model the instance-dependent preference flipping incurred during preference data annotation?}
\end{center}

In response, we propose a statistically consistent approach to align with corrupted human feedback, which we refer to as Flipping-Aware Direct Preference Optimization (FA-DPO). 
This approach ensures that, given accurate estimation of the flipping model, the learned policy will achieve consistency comparable to learning the policy under clean data.
We first explicitly model the flipping probability on the basis of the Bradley-Terry model, to connect with the posterior probabilities of the observed label. 

Specifically, the annotation process under our framework can be viewed as two sequential stages: \ding{182}~labeling according to true human intention and \ding{183}~instance-dependent label contamination, where the instance-dependent preference flipping occurs after the labeling stage~\ding{182}, representing a transition from the true label to the flipped label for each sample. 
In this way, given a corrupted human preference dataset, FA-DPO can post-train the LLM parameters robustly and produce optimal policies, as with the true labels, via correction for the original BT model loss with the posterior of the observed corrupted labels.


In particular, we further introduce an iterative optimization framework that jointly optimizes both the flipping estimation model and the LLM in post-training.
To capture the transition between the original label and the observed label, 
we leverage a classification module on relevant features to construct the flipping probability within a preference pair. 
The main contributions of this paper are three-fold:
\begin{enumerate}
    \item We investigate a challenging RLHF / DPO setting with instance-dependent preference flipping, 
    and propose a novel probabilistic model to characterize the noise. 
    \item We address the instance-dependent estimation of preference flipping probabilities via a classification model that incorporates informative preference features studied in natural language processing.
    \item We propose a simple yet efficient iterative optimization algorithm based on RLHF and DPO to capture the noise in the offline dataset during post-training. 
    The whole training pipeline is validated in our experiments.
\end{enumerate}

\section{Related Works}
We review the previous works on preference alignment and (generalized) robust RLHF in this section. Due to context limitation, we leave the complete review to Appendix B for the reader's convenience.

\paragraph{Preference alignment}
The most well-known approach for preference alignment is Reinforcement Learning from Human Feedback~\citep[RLHF]{ziegler2019fine, ouyang2022training}, which involves training a \emph{reward model} to capture human preferences and then guiding LLMs to \emph{generate high-reward responses using reinforcement learning algorithms} such as Proximal Policy Optimization (PPO) \citep{schulman2017proximal}. 
However, in practice RL-based methods can be complex and unstable during training~\citep{rafailov2024direct, wu2024fine, yuan2023rrhf}. As a result, recent research has focused on simpler and more stable alternatives to RLHF, namely, direct preference alignment \citep{rafailov2024direct, zhao2023slic, ethayarajh2024kto, azar2024general, meng2024simpo}.

\paragraph{RLHF against perturbations}
Current approaches of robust RLHF can be categorized into three main types. 
\ding{182}~Noise fitting \citep{bukharin2024robust} involves making assumptions on the noise in the data and incorporating this modeling into the reward learning process, which will be jointly optimized with the parameterized reward function.
\ding{183}~Sample selection (or sample re-weighting) methods \citep{cheng2024rime} leverage this phenomenon to identify clean samples based on the loss values observed during the training process. 
Some approaches address corruption in input data through \ding{184}~robust loss design \citep{chowdhury2024provably, liang2024robust}, focusing on constructing loss functions that are more resistant to data noise; this technique is also known as label smoothing~\citep{mitchell2023note}.

\section{Preliminaries}
\label{sec: preliminary}

In this section, we introduce the basics of RLHF for LLM alignment in advance of the formal proposal of our methodology in \Cref{sec: method}.

\subsection{RLHF for LLM alignment}

Let an input (prompt) of an LLM be $x \in \mathcal{X}$ and the generated output (response) be $y \in \mathcal{Y}$.
In the formulation for RLHF~\citep{rafailov2024direct}, the LLM is viewed as a policy $\pi_\theta(y\mid x)$ parameterized by $\theta$, which outputs an action $y$ (response) based on the state $x$ (prompt).
Preference data is further collected and annotated by human labelers, denoted as $y_w\succ y_l\mid x$, where $y_w$ is the preferred response and $y_l$ is the dispreferred one in $(y_1,y_2)$ for the prompt $x$.


The pre-trained model first undergoes on round of supervised fine-tuning (SFT), resulting in a reference model $\pi_\text{ref}$.
Then the Bradley-Terry (BT) model~\citep{bradley1952rank} is employed to relate the preference data $\{y_w^i \succ y_l^i\}$ to a reward model $r(x, y)$. 
The connection is formulated as:
\begin{equation}
\label{eq:Bradley-Terry model}
p^*(y_w \succ y_l \mid x) = \sigma \big(r^*(x, y_w) - r^*(x, y_l)\big),
\end{equation}
where $\sigma$ is the standard sigmoid function, and $r^*(\cdot)$ is the optimal reward model. 
Using the BT model and the maximum likelihood principle, the loss for learning the reward model is:
\begin{equation}
\begin{aligned}
\label{eq: reward modeling}
\mathcal{L}_R(\phi) = -\mathbb{E}_{(x,y_w,y_l)\sim\mathcal{D}}\left[\log \sigma\big(r_\phi(x, y_w) - r_\phi(x, y_l)\big)\right],
\end{aligned}
\end{equation}
where $r_\phi(\cdot)$ is the reward model parameterized by $\phi$.

After training the reward model $r_\phi(\cdot)$ from \Cref{eq: reward modeling}, reinforcement learning (RL) is applied to optimize the LLM $\pi_\theta$ with the reward signals provided by $r_\phi(\cdot)$.
The optimization objective for $\pi_\theta$ is formulated as:
\begin{equation}
\begin{aligned}
\label{eq:policy optimization}
\mathcal{L}_\pi(\theta) = & -\mathbb{E}_{x\sim \mathbb{P}_x, y \sim \pi_\theta(\cdot|x)} \left[r_\phi(x, y)\right] \\ 
& + \beta \cdot \mathbb{E}_{x\sim \mathbb{P}_x} \left[\mathbb{D}_\mathrm{KL} \left(\pi_\theta(\cdot\mid x) || \pi_\text{ref}(\cdot\mid x)\right)\right],
\end{aligned}
\end{equation}
where $\mathbb{P}_x$ represents the marginal distribution of the prompt $x$. 
The first term corresponds to reward maximization in standard RL optimization. 
The second term is a KL divergence that constrains the update of $\pi_\theta$ to not deviate from the reference model $\pi_\text{ref}$, 
$\beta$ is the coefficient that weights the KL divergence between $\pi_\theta$ and $\pi_{\text{ref}}$.

\subsection{DPO as efficient RLHF}

Direct preference optimization (DPO) is an algorithm proposed for practical efficiency (compared to original RLHF), that merges the two stages in RLHF, reward modeling and policy optimization, into a single step. 
In DPO, the policy is optimized directly using the offline dataset~$\mathcal{D}$ without constructing an explicit reward model.

As shown by \citet{peng2019advantage}, the closed-form solution of the conditional distribution $\pi(y \mid x)$ that minimizes \Cref{eq:policy optimization} is:
\begin{equation}
\label{eqn:DPO-sol}
\pi_r(y \mid x) = \frac{1}{Z(x)}\pi_\text{ref}(y \mid x) \exp\left(\frac{1}{\beta} r_\phi(x, y)\right),
\end{equation}
where $Z(x) = \sum_y \pi_\text{ref}(y \mid x) \exp\big({1}/{\beta} \cdot r_\phi(x, y)\big)$ is the partition function for normalization.
By applying this result to \Cref{eq: reward modeling}, the DPO loss function that includes only the parameterized $\pi_\theta$ as the optimization variable is derived as:
\begin{equation}
\label{eq: DPO loss}
\mathcal{L}_\text{DPO}(\theta) = -\mathbb{E}_{(x,y_w,y_l)\sim\mathcal{D}}\left[\log\sigma\left(
\hat{r}_\theta(x,y_w) - \hat{r}_\theta(x,y_l)
\right)\right],
\end{equation}
where $\hat{r}_\theta(x,y)=\beta\log \big({\pi_\theta(y|x)}/{\pi_\text{ref}(y|x)} \big)$ is the implicit reward model derived from $\pi_\theta$.

\section{Method}
\label{sec: method}



In this section, we begin by providing an overview of the motivation 
to model 
instance-dependent preference flipping. 
We then formulate the flipping procedure under the RLHF framework, 
and illustrate how this estimation can be integrated into the standard RLHF pipeline. 
Finally, we present a detailed model design for the instance-dependent preference flipping probability estimation in FA-DPO.

\subsection{Motivations}


It is widely accepted that in real-world scenarios, the intrinsic preference labeling mechanism of humans can be unified as a general human intention model, represented by the Bradley-Terry model~\citep{bai2022training}. 
Noisy human feedback, however, usually stems from external factors rather than inherent errors in human decisions. 
For instance, environmental distractions may compromise an annotator's focus, reducing labeling accuracy, and 
external noise can also be introduced by maliciously altering original annotations.

Consequently, this corruption process can be viewed as a post-transition applied to the initial labeling mechanism governed by the human intention model. 
We model this process as instance-dependent preference flipping, 
positing that the probability of a flip is correlated with the data content as well as the relation between each pair of responses.

We dig into the statistical structure of the flipping pattern in the preference data as opposed to the noise sparsity assumption in \citep{bukharin2024robust}.
The proposed modeling, as stated in \Cref{sec:flip}, is supposed to exploit the relation between the noisy preference posterior and the true likelihoods.
The neural network module to capture the noise is introduced in \Cref{sec:noise-modeling}.

\subsection{Flipping-aware Loss}
\label{sec:flip}

We start the derivation of the flipping-aware loss from the RLHF model, and then extend the result to the DPO setting. 
Given a corrupted dataset $\tilde{\mathcal{D}}$, for each noisy triplet $(x,\tilde{y}_w,\tilde{y}_l)$, we denote the sample as $\tilde{\bm{x}}$ for simplicity. 
Following the principle of maximum likelihood, the usual loss for preference modeling is:
\begin{equation}
    \label{eq:loss correction}
    \mathcal{L}_\text{MLE}=-\mathbb{E}_{(x,\tilde{y}_w,\tilde{y}_l)\sim\tilde{\mathcal{D}}}
    [\log{\mathbb{P}}\{\tilde{y}_w\succ\tilde{y}_l\mid x\}].
\end{equation}
In standard RLHF, we only have the parameterized preference probability under clean data, i.e., $\mathbb{P}\{y_w\succ y_l\mid x\}$;
the direct usage of the standard RLHF loss in \Cref{eq:loss correction} can be sub-optimal. 
To ease the discussion of how to bridge the gap between preference distributions under clean and noisy data, we first formalize the preference flipping process through the following proposition:
\begin{proposition}[Instance-dependent preference flipping]
    \label{def:preference flipping}
    For any input $\tilde{\bm{x}}$, the corrupted preference probability, under the instance-dependent preference flipping setting, relates to the true preference likelihood via:
    \begin{equation*}
        \tilde{\mathbb{P}}\{\tilde{y}_w\succ\tilde{y}_l\mid x\} = (1-\varepsilon_{\tilde{\bm{x}}})p + \varepsilon_{\tilde{\bm{x}}}(1-p),
    \end{equation*}
    where $\varepsilon_{\tilde{\bm{x}}}$ represents the instance-specific flipping probability for triplet $\tilde{\bm{x}}= (x,\tilde{y}_w,\tilde{y}_l)$,
    and $p$ denotes the true likelihood $\mathbb{P}\{\tilde{y}_w\succ \tilde{y}_l\mid x\}$ for brevity.
\end{proposition}
The proof of \Cref{def:preference flipping} is direct and omitted. We can then establish the relation between the corrupted posterior $\tilde{\mathbb{P}}\{\tilde{y}_w \succ \tilde{y}_l \mid x\}$ observed in noisy data and the underlying clean probability $\mathbb{P}\{\tilde{y}_w \succ \tilde{y}_l \mid x\}$. 
This enables us to recover the true preference probabilities by accounting for the instance-dependent flipping process:
\begin{equation}
    \label{eq:flipo}
    \mathcal{L}_\text{FA-DPO} = -\mathbb{E}_{\tilde{\bm{x}}\sim\tilde{\mathcal{D}}}\left[\log\left((1-\varepsilon_{\tilde{\bm{x}}})p + \varepsilon_{\tilde{\bm{x}}}(1-p)\right)\right],
\end{equation}
which reads the loss design in FA-DPO.
For practical implementation with the BT model in reward-based RLHF, we parameterize $p$ with $\phi$ and denote:
\begin{equation*}
    p_\phi = \sigma(r_\phi(x,\tilde{y}_w) - r_\phi(x,\tilde{y}_l)).
\end{equation*}
Similarly, for the DPO parameterization, the preference probability is given by substituting the probabilistic modeling \Cref{eqn:DPO-sol} into the formula above:
\begin{equation*}
    p_\theta = \sigma\left(\beta\log\frac{\pi_\theta(\tilde{y}_w\mid x)}{\pi_\text{ref}(\tilde{y}_w\mid x)} - \beta\log\frac{\pi_\theta(\tilde{y}_l\mid x)}{\pi_\text{ref}(\tilde{y}_l\mid x)}\right).
\end{equation*}

\paragraph{Comparison with cDPO and rDPO.} 
To better understand FA-DPO, we compare it with related approaches, cDPO~\citep{mitchell2023note} and rDPO~\citep{chowdhury2024provably}, from the perspective of gradients.

First, we recall standard DPO directly substitutes the paramterized preference probability $\mathbb{P}_\theta\{\tilde{y}_w \succ \tilde{y}_l \mid x\}$ for the corrupted posterior $\tilde{\mathbb{P}}\{\tilde{y}_w \succ \tilde{y}_l \mid x\}$, 
ignoring the preference flipping mechanism underlying the corrupted datasets.
Other robust losses, such as cDPO~\citep{mitchell2023note}, instead similarly follow the principle of maximum likelihood and adopt a loss correction technique, 
but 
the $\varepsilon$ parameter in cDPO is a hyperparameter irrelevant to each sample
(while our loss 
considers an instance-dependent preference flipping setting).
On the basis of cDPO, rDPO~\citep{chowdhury2024provably} further debias the loss function.

To better understand the difference between these methods, 
we compare the gradient weights of these methods with that of FA-DPO.

Following the notations in~\citet{chowdhury2024provably}, we characterize the gradient of DPO-like methods as
\small
\begin{equation}
    \nabla_\theta\mathcal{L} = -\zeta \cdot \beta\left( 
    \nabla_\theta\log\pi_\theta(\tilde{y}_w\mid x) - \nabla_\theta\log\pi_\theta(\tilde{y}_l\mid x) 
    \right).
\end{equation}
\normalsize

The weighting coefficient $\zeta$ for each method can be related to the base DPO weight coefficient $\zeta_{\text{DPO}} = 1 - p_{\theta}$, where cDPO applies a small reduction $\zeta_{\text{cDPO}} = \zeta_{\text{DPO}} - \varepsilon$ while rDPO introduces an additive correction $\zeta_{\text{rDPO}} = \zeta_{\text{DPO}} + \frac{\varepsilon}{1-2\varepsilon}$ to adjust the weighting behaviors, respectively.
We specify the coefficient $\zeta_\text{FA-DPO}$ in FA-DPO as follows.
The derivation is deferred in Appendix A.

\begin{lemma}[Gradient weight coefficient]
\label{lemma:gradient weights}
For a triplet $\tilde{\bm{x}} = (x,\tilde{y}_w,\tilde{y}_l)$, we have the gradient weight coefficient for FA-DPO as
    \begin{equation*}
        \zeta_\text{FA-DPO}=\frac{(1-2\varepsilon_{\tilde{\bm{x}}})p_\theta}{(1-2\varepsilon_{\tilde{\bm{x}}})p_\theta+\varepsilon_{\tilde{\bm{x}}}}\cdot\zeta_\text{DPO}.
    \end{equation*}
\end{lemma}

The above lemma demonstrates that our weighting scheme constitutes a reparametrization of the DPO gradient weight, distinct from the additive correction approaches employed by cDPO and rDPO.
Crucially, when no flipping occurs ($\varepsilon_{\tilde{\mathbf{x}}} = 0$), the gradient reduces exactly to that of the standard DPO. 
For low flipping probabilities ($\varepsilon_{\tilde{\mathbf{x}}} < 0.5$), the weight increases with the model's confidence ($p_\theta$), enhancing stability during convergence and improving robustness to noise compared to the fixed correction mechanisms of cDPO and rDPO. 
At a flipping probability of 0.5, indicating inherent ambiguity in the preference signal, the weight becomes zero, automatically filtering out these low-margin samples that could otherwise impair model performance. 
Most significantly, when $\varepsilon_{\tilde{\mathbf{x}}} > 0.5$, the gradient direction reverses to optimize $\mathbb{P}\{\tilde{y}_l \succ \tilde{y}_w \mid x\}$ instead of $\mathbb{P}\{\tilde{y}_w \succ \tilde{y}_l \mid x\}$ under conditions of high model uncertainty. This demonstrates self-correction for samples with high detected flipping rates---a capability absent in cDPO and rDPO which apply uniform corrections. Furthermore, the weight approaches zero when high model confidence contradicts high estimated $\varepsilon_{\tilde{\mathbf{x}}}$, yielding robust weights that are jointly determined by both the flipping probability and the policy model's confidence.

\subsection{Transition Probability Modeling}
\label{sec:noise-modeling}

\paragraph{Preference flipping modeling}
To overcome the limitations of existing robust DPO approaches that assume either (1) a uniform flipping ratio across all samples or (2) sparse noise patterns within a given dataset, we introduce an instance-dependent preference flipping module that dynamically estimates sample-specific flipping probabilities based on instance features.
we model this probability as a logistic regression function of input-dependent features:
\begin{equation}
    \varepsilon_{\tilde{\bm{x}}} = \sigma(\langle\omega,h(\tilde{\bm{x}})\rangle + \omega_0),
    \label{eq:noise_model}
\end{equation}
where $h:\mathcal{X}\rightarrow\mathbb{R}^d$ is the feature map and $\omega\in \mathbb{R}^d$ are learnable parameters. 

\paragraph{Feature map construction}
We deliberately design the feature map $h(\cdot)$ to incorporate three concepts validated to be effective in language modeling.
Notably, the features are supposed to be permutation-equivariant to the response pairs, since the order thereof can be arbitrary in the corrupted sample triplet $\tilde{\bm{x}} = (x,\tilde{y}_w,\tilde{y}_l)$.

~\\
\noindent\emph{Response Length.}
It is noted that longer responses increase cognitive load for human annotator, which may increase the error rate~\citep{chen2024humans}. 
We compute both the average lengths and the length difference within a preference sample triplet $(x,\tilde{y}_w,\tilde{y}_l)$ as the first feature,
\begin{equation*}
    h_\text{len}(\tilde{\bm{x}}) = \left[\frac{|\tilde{y}_w|+|\tilde{y}_l|}{2}, \left||\tilde{y}_w|-|\tilde{y}_l|\right|\right]^\top.
\end{equation*}
Here, even we change the order of the response pair to $(x,\tilde{y}_l,\tilde{y}_w)$,
the feature map $h_\text{len}(\cdot)$ is invariant.

~\\
\noindent\emph{Perplexity (PPL).}
Perplexity reflects the uncertainty or complexity of a probability distribution. 
High PPL often correlates with sentences that are hard for humans to comprehend, thereby increasing annotation difficulty~\citep{kong2024perplexity}.
\begin{equation*}
    h_{\text{ppl}}(\tilde{\bm{x}}) = \left[\frac{\log\left( \pi_\theta(\tilde{y}_w|x)\pi_\theta(\tilde{y}_l|x)\right)}{2},\left|\log \frac{\pi_\theta(\tilde{y}_w|x)}{\pi_\theta(\tilde{y}_l|x)}\right|\right]^\top.
\end{equation*}

~\\
\noindent\emph{Reward Margin.}
The reward margin implicitly quantifies the model's confidence in distinguishing preferred from dispreferred responses, which is often used for data selection in preference learning~\citep{wu2024beta, huang2025larger}.
The corresponding feature map is
\small
\begin{equation*}
    h_{\text{margin}}(\tilde{\bm{x}}) = \left[\frac{\hat{r}_\theta(x,\tilde{y}_w) + \hat{r}_\theta(x,\tilde{y}_l)}{2},\left|\hat{r}_\theta(x,\tilde{y}_w) - \hat{r}_\theta(x,\tilde{y}_l)\right|\right]^\top
\end{equation*}
\normalsize
where $\hat{r}_\theta(x,y)=\beta\log\frac{\pi_\theta(y\mid x)}{\pi_\text{ref}(y\mid x)}$ is the implicit reward function induced by DPO.

These features are then concatenated, along with a scalar~$1$ (for the bias term in $\omega$), and scaled to form the final feature map: 
\begin{equation}
    h(\tilde{\bm{x}}) = \left[h_\text{len}(\tilde{\bm{x}}),h_\text{ppl}(\tilde{\bm{x}}),h_\text{margin}(\tilde{\bm{x}})\right]^\top
\end{equation}

\paragraph{Iterative update}
To optimize both the preference flipping model and the LLM, 
we design an iterative update paradigm, which is also adopted in previous work for robust RLHF~\citep{bukharin2024robust}.

Regarding the concrete design,
empirical studies have demonstrated that deep neural networks exhibit a consistent learning trajectory, initially capturing generalizable patterns before eventually overfitting to noisy training instances~\citep{cheng2024rime}.
To adapt to this characteristic, in practice we utilize the learned capability of the LLM during the initial stage of training to optimize the flipping model first 
(this operation is referred to as \emph{warmup}),
and then we iteratively update the two models.
The complete training algorithm is presented in Algorithm 1 in Appendix D.

\subsection{Theoretical Analysis}

This section establishes the theoretical foundations of our approach. 
Specifically, we demonstrate that the proposed loss function for preference flipping yields desirable statistical properties: 
consistency under noise and convergence guarantees for 
the flipping model parameters.

We first present the key result that the minimizer of our FA-DPO loss function, operating on the corrupted (flipped preference) data distribution $\tilde{\mathcal{D}}$, coincides with the minimizer of the original loss function on the underlying clean data distribution $\mathcal{D}$ in RLHF. 
This \emph{consistency} property guarantees that, asymptotically, our method recovers the same optimal model parameters as would be obtained if trained on clean preference data. We provide the detailed proof in Appendix A.

\begin{theorem}[Consistency of $\bm{p}_\theta$]
    \label{thm:consistency}
    Given both the corrupted preference data distribution $\tilde{\mathcal{D}}$ induced by the flipping process and the unobserved clean preference data distribution $\mathcal{D}$, the following equality holds:
    \begin{equation*}
        \argmin_\phi -\mathbb{E}_{\tilde{\bm{x}}\sim\tilde{\mathcal{D}}}[\log\tilde{p}_\phi]=
        \argmin_\phi -\mathbb{E}_{\bm{x}\sim\mathcal{D}}[\log{p}_\phi],
    \end{equation*}
    where $\tilde{p}_\phi = (1 - \varepsilon_{\tilde{\bm{x}}}) p_\phi + \varepsilon_{\tilde{\bm{x}}} (1 - p_\phi)$ represents the predicted probability under the flipping model. 
\end{theorem}

This result holds under the specific parameterization where the flipping noise is marginalized via $\tilde{p}_\phi$.
Furthermore, the core consistency principle holds equivalently under the DPO parameterization:
replacing $p_\phi$ with $p_\theta$ and $\tilde{p}_\phi$ with $\tilde{p}_\theta$, 
the same result applies to the policy $\theta$ optimized via the DPO loss adjusted for flipping.

On the other hand, through a coordinate descent perspective on the iterative updates, the optimization of the flipping model reduces to the convex logistics regression.
We present the following convergence result 
for self-containedness.

We first make the technical assumption that the features $h(\tilde{\bm{x}})$ and the parameters $\omega$ for the preference flipping model are bounded, i.e., $\|h(\tilde{\bm{x}})\|\leq B$, $\|\omega\|\leq B_\omega$.
Then, we further assume that the constructed features satisfy the following coverage assumption, 
which is a common assumption in robust machine learning and the 
implied feature diversity is critical to prevent collapse or degenerate solutions.

\begin{assumption}[Feature coverage]
    \label{assump:coverage}
    Given the corrupted data distribution $\tilde{\mathcal{D}}$ and the feature map $h: \mathcal{X} \to \mathbb{R}^d$, the population covariance matrix of features satisfies
    \begin{equation*}
        \lambda_{\min}\left( \mathbb{E}_{\tilde{\bm{x}}\sim\tilde{\mathcal{D}}}\left[h(\tilde{\bm{x}})h(\tilde{\bm{x}})^\top\right] \right) > 0.
    \end{equation*}
\end{assumption}

This assumption implies that the feature vectors are not concentrated in a low-dimensional subspace,
which eliminates potential collinearity. 
The parameters $\omega$ are thus identifiable from the data. 

Under the boundedness assumption, Assumption \ref{assump:coverage}, and the condition that the reward or policy model provides accurate predictions ($p_\phi = p^*$ or $p_\theta = p^*$), we can establish fast convergence for the estimator of the flipping parameters $\omega$ obtained via gradient descent:

\begin{theorem}[Linear Convergence of $\hat{\omega}$]
    \label{thm:convergence for noise}
    Given loss function in \Cref{eq:flipo}, if $p_\phi=p^*$ or $p_\theta=p^*$ and
    for the gradient descent update with step size $\eta > 0$:
    \begin{equation*}
        \omega^{(t+1)} = \omega^{(t)} - \eta \nabla_\omega \mathcal{L}_\text{FA-DPO}(\omega^{(t)}),
    \end{equation*}
    the sequence of parameter estimates $\{\omega^{(t)}\}$ converges \emph{Q-linearly}  to the optimal parameter $\omega^*$:
    \begin{equation*}
        \|\omega^{(t+1)} - \omega^*\|^2 \leq (1 - \eta \mu) \|\omega^{(t)} - \omega^*\|^2,
    \end{equation*}
    for some $\mu>0$, the convergence holds when $0<\eta<\frac{2}{L}$, where $L$ is the smoothness constant for $\mathcal{L}_\text{FA-DPO}$.
\end{theorem}

\Cref{thm:convergence for noise} provides the important guarantee that, when initialized with an accurate reward model or policy (reflecting the true clean preference probability $p^*$), the gradient descent update on the preference flipping model parameters $\omega$ converges rapidly at a linear rate. 
Consequently, this justifies the iterative procedures used in our algorithm.

\section{Experiments}

We investigate the effectiveness of FA-DPO mainly on aligning the LLMs with preference pairs under the instance-dependent preference flipping setting.

\begin{table*}[t!]
    \small
    \centering
    \begin{tabular}{llccccccccc}
    \toprule
     \multicolumn{11}{c}{\textbf{Ultrafeedback} (61.1k)} \\
     \midrule
    $\eta$ & \multicolumn{2}{c}{0\%} & \multicolumn{2}{c}{10\%} & \multicolumn{2}{c}{20\%} & \multicolumn{2}{c}{30\%} & \multicolumn{2}{c}{40\%} \\
    Methods & Acc $\uparrow$ & WR $\uparrow$ & Acc $\uparrow$ & WR $\uparrow$ & Acc $\uparrow$ & WR $\uparrow$ & Acc $\uparrow$ & WR $\uparrow$ & Acc $\uparrow$ & WR $\uparrow$ \\
    \midrule
    DPO & 68.22 & \textbf{67.00} & 61.77 & 60.10 & 58.63 & 56.35 & 55.43 & 64.70 & 51.87 & 64.35 \\
    SIMPO & 63.64 & 62.60 & 63.73 & 51.20 & 57.81 & 56.50 & 55.53 & 54.90 & 51.77 & 55.80 \\
    ROPO & \underline{70.75} & 66.80 & 64.93 & 66.85 & 58.93 & 65.55 & 55.23 & 64.40 & \underline{54.07} & \underline{67.25} \\
    cDPO & 67.20 & 66.65 & 62.57 & \underline{67.80} & 59.00 & \underline{66.05} & 56.67 & \underline{66.55} & 53.93 & 67.05 \\
    rDPO & 70.13 & 65.60 & \underline{65.90} & 56.65 & \underline{61.70} & 54.85 & \underline{56.87} & 57.85 & 47.67 & 57.80  \\
    FA-DPO & \textbf{73.05} & \underline{66.85} & \textbf{67.20} & \textbf{68.45} & \textbf{69.77} & \textbf{66.90} & \textbf{70.97} & \textbf{69.80} & \textbf{70.77} & \textbf{69.80} \\
    (\#Improv.)  & +2.30 & -0.15 & +1.30 & +0.65 & +8.07 & +0.85 & +14.10 & +3.25 & +16.70 & +2.55 \\
    \midrule
    \multicolumn{11}{c}{\textbf{HH\_Golden} (42.5k)} \\
    \midrule
    DPO & 98.89 & 87.00 & 93.50 & 34.05 & 83.53 & 29.95 & 73.30 & 21.05 & 58.63 & 26.85 \\
    SIMPO & 98.27 & 65.10 & 93.63 & 29.95 & 82.83 & 24.90 & 73.33 & 22.40 & 60.93 & 27.50\\
    ROPO & 98.80 & 70.50 & 93.27 & 81.15 & 82.87 & 54.30 & 72.33 & 53.50 & \underline{62.00} & 49.20\\
    cDPO & \underline{99.24} & \underline{88.35} & 93.20 & 56.95 & 82.50 & \underline{62.30} & 73.30 & 37.75 & 60.87 & 49.05\\
    rDPO & 96.80 & 77.30 & \underline{96.67} & \underline{82.80} & \underline{87.83} & 53.55 & \underline{73.73} & \underline{58.10} & 56.80 & \underline{54.10}\\
    FA-DPO & \textbf{99.61} & \textbf{88.70} & \textbf{99.17} & \textbf{88.90} & \textbf{98.10} & \textbf{82.85} & \textbf{99.02} & \textbf{87.70} & \textbf{98.83} & \textbf{78.60} \\
    (\#Improv.) & +0.37 & +0.35 & +2.50 & +6.10 & +10.27 & +20.55 & +25.29 & +29.60 & +36.83 & +24.50 \\
    \bottomrule
    \end{tabular}
    \caption{
    Model performance under different preference flipping ratios (0\%-40\%) for Pythia-1B on Ultrafeedback and HH\_Golden.
    All values represent percentages, with \textbf{bold} indicating the highest score per column and \underline{underline} denoting the runner-up. The (\#Improv.) row quantifies absolute performance gain over the runner-up baseline.
    }
    \normalsize
\label{tab:pythia1b performance}
\end{table*}

\subsection{Experiment Setup}

\label{sec:exp-setup}

\paragraph{Datasets and models}
We conduct experiments mainly on two preference datasets: UltraFeedback~\citep{cui2024ultrafeedback} and Anthropic's HH\_Golden~\citep{Anthropic2024hhGolden}. 
For UltraFeedback, we follow the spirit of \citet{chen2025compo} and first filter out samples with low score margins ($\leq$ 0.5) to obtain a cleaner subset for noise injection. This results in 42.4k/61.1k training samples and 1.4k/2k test samples. However, in experiments without manual label flipping, we retain the full training set while using the filtered test set for evaluation. 
For the backbone LLM models, we first test Pythia-1B on various levels of noise, 
and then we scale our method on larger models, LLama-3.1-8B and Mistral-7B, to validate the model's generation capabilities under data contamination.
The detailed training setup are listed in Appendix E.

\paragraph{Instance-dependent flipping}
To validate our designs, we construct datasets with simulated instance-dependent preference flipping based on the following steps:
\begin{enumerate}
    \item Randomly initialize a preference flipping model $\mathcal{N}_\varepsilon(\vartheta)$ (logistic model) on data features discussed in \Cref{sec:noise-modeling}, and the model outputs a flip probability $\varepsilon_{\tilde{\bm{x}}}$ for the current preference pair.
    \item To determine whether to flip a preference sample, we set a threshold $\tau$ such that if the flipping probability $\varepsilon_{\tilde{\bm{x}}}$ computed by the noise model is high enough to exceed this threshold, we would flip the preference. This is to maintain controlled stochasticity during preference flipping. During our experiment, we choose $\tau=0.8$.
    \item To control the total ratio of flips, we train the parameters $\vartheta$ on the training set $\mathcal{D}_\text{train}$ using the following loss function during noise model initialization, which forces the quantile of the flipping probability in the interval $[\tau, 1]$ to be around a pre-determined flipping ratio $\eta$.
    $$
        \mathcal{L}_\varepsilon=\mathbb{E}_{(x,y_w,y_l)\sim \mathcal{D}_\text{train}}\left[\eta-\frac{\mathbb{I}\left(\mathcal{N}_\varepsilon(x,y_w,y_l;\vartheta)\geq\tau\right)}{|\mathcal{D}_\text{train}|}\right]^2.
    $$
\end{enumerate}

After training on the contaminated training set, 
we test the model's performance on the clean test set which has no noise injected.
For experiments with simulated flipping ($\eta>0$), we intentionally restrict the capability of preference flipping model via only using length-based features,
recovering the practical mismatch between estimations and real-world mechanisms.
For clean datasets without added flipping ($\eta=0$), we utilize the full feature set to model the original dataset's inherent noise patterns.

\paragraph{Evaluation metrics}

We use \emph{prediction accuracy} (ACC) and \emph{win rate} (WR) as our evaluation metrics. 

In particular, we evaluate each model's \emph{prediction accuracy} on a clean test set $\mathcal{D}_\text{test}$ across different noise levels. 
The accuracy is computed by comparing the predicted rewards for ``chosen'' versus ``rejected'' responses:
\begin{equation*}
    \mathrm{Acc} = \mathbb{E}_{(x,y_w,y_l)\sim\mathcal{D}_{\text{test}}} \left[\mathbb{I} \left(\hat{r}_\theta(x, y_w) > \hat{r}_\theta(x, y_l) \right)\right],
\end{equation*}
where $\mathbb{I}$ is the indicator function, and $\hat{r}_\theta$ is the implicit reward model induced by DPO.

To compute the \emph{win rate}, we employ different evaluators based on model capabilities: DeepSeek-V3 for Pythia-1B, and GPT-4o for both LLama-3.1-8B and Mistral-7B. 
The calculation is performed as follows:
\begin{equation*}
    \mathrm{WR}=\frac{\#\text{(Win)}+\#\text{(Tie)}/2}{\#\text{(Comparisons)}},
\end{equation*}
where $\#\text{(Win)}$ and $\#\text{(Tie)}$ represent the number of wins and ties compared to the \emph{reference model} (In our case, the SFT model), respectively, and $\#\text{(Comparisons)}$ denotes the total number of comparisons between the two models.

For \emph{prediction accuracy}, we compute 5 runs for each method and report the mean value, with each run extracting randomly 1k samples from $\mathcal{D}_\text{test}$ for testing.
For \emph{win rate}, we extract 1k prompts from $\mathcal{D}_\text{test}$, and generate 1k responses from the trained policy model and the corresponding SFT model, respectively. 
We refer the readers to Appendix E for the detailed prompt template of LLM evaluators.

\begin{figure*}[t!]
    \centering
    \begin{subfigure}{0.3\textwidth}
        \includegraphics[width=\linewidth]{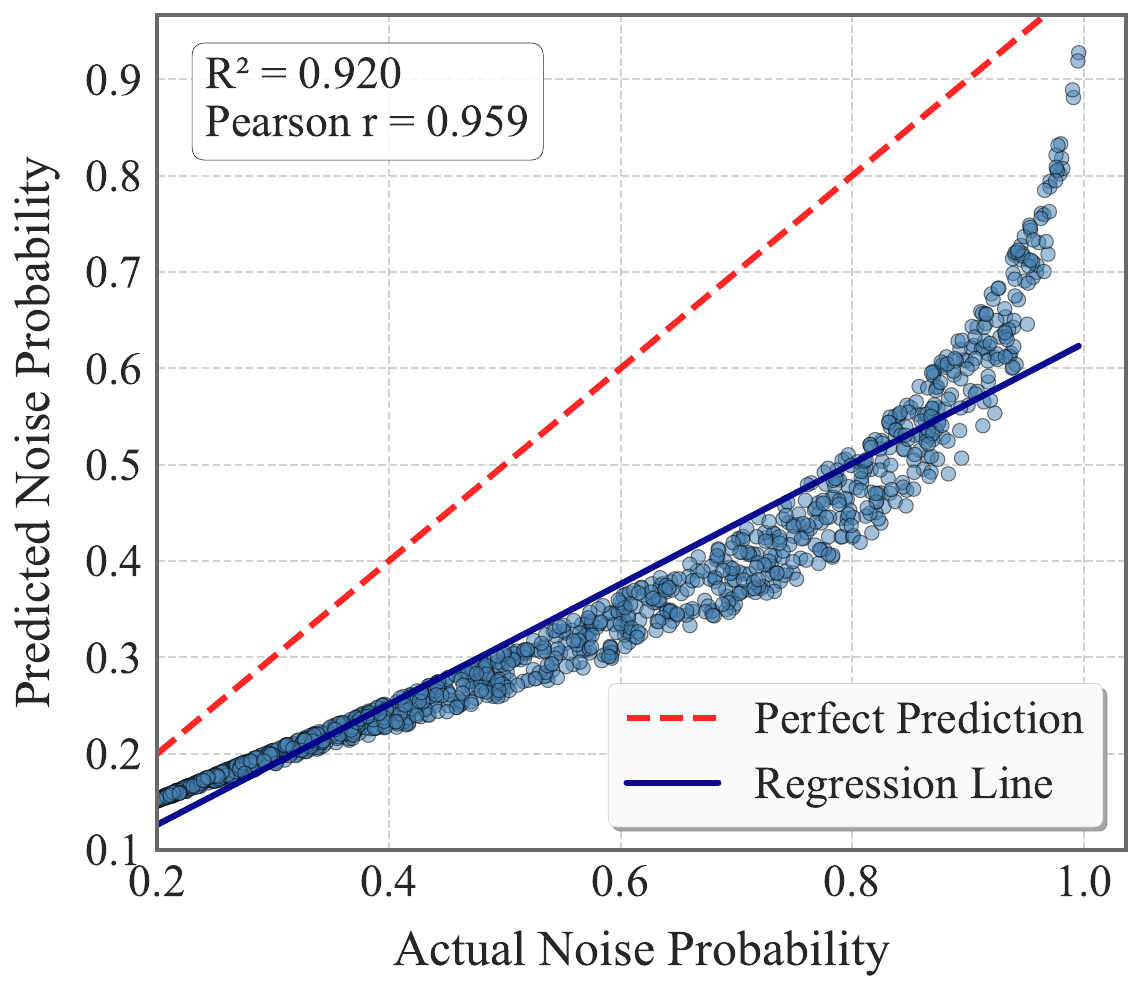}
        \caption{Prediction correlation}
    \end{subfigure}
    \hfill
    \begin{subfigure}{0.3\textwidth}
        \includegraphics[width=\linewidth]{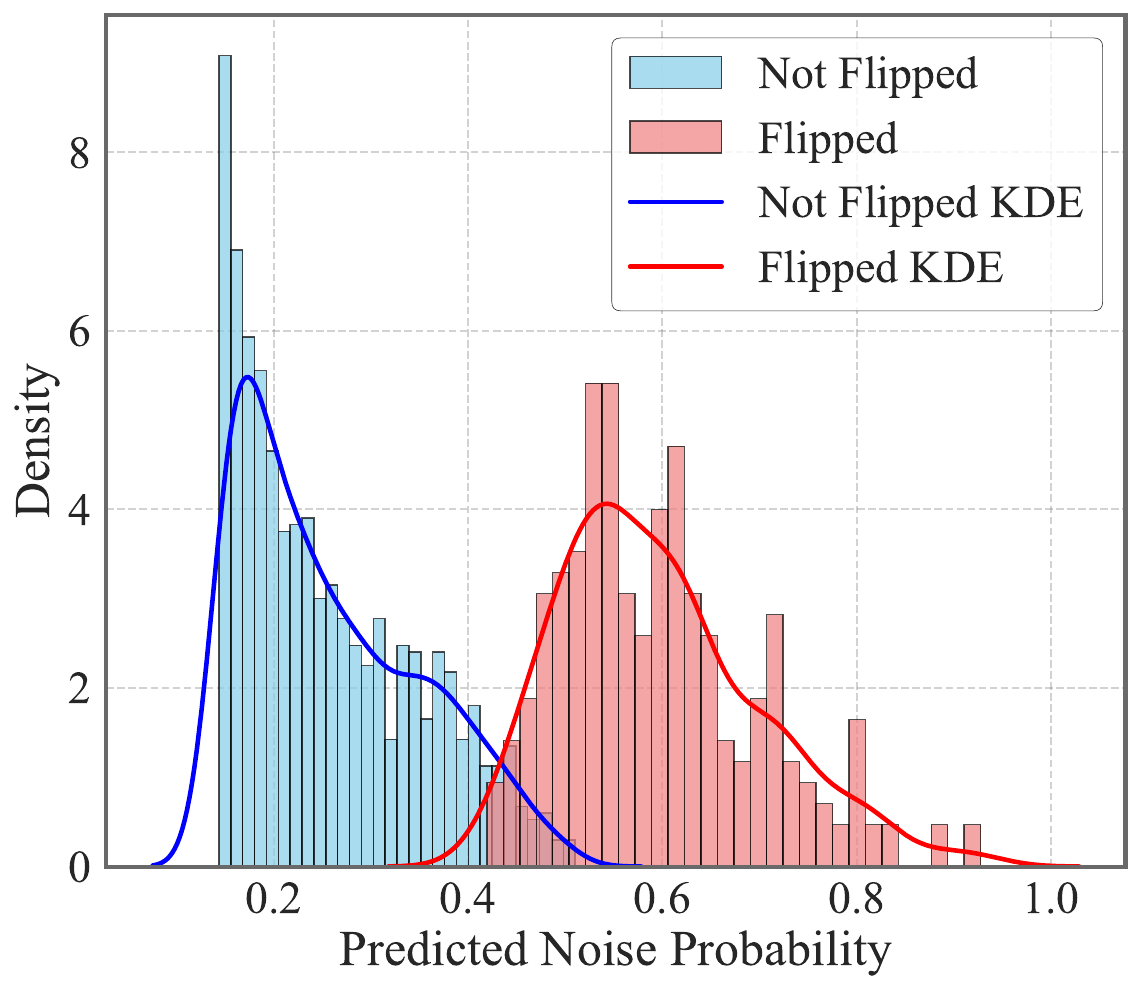}
        \caption{Noise distribution by flipping status}
    \end{subfigure}
    \hfill
    \begin{subfigure}{0.3\textwidth}
        \includegraphics[width=\linewidth]{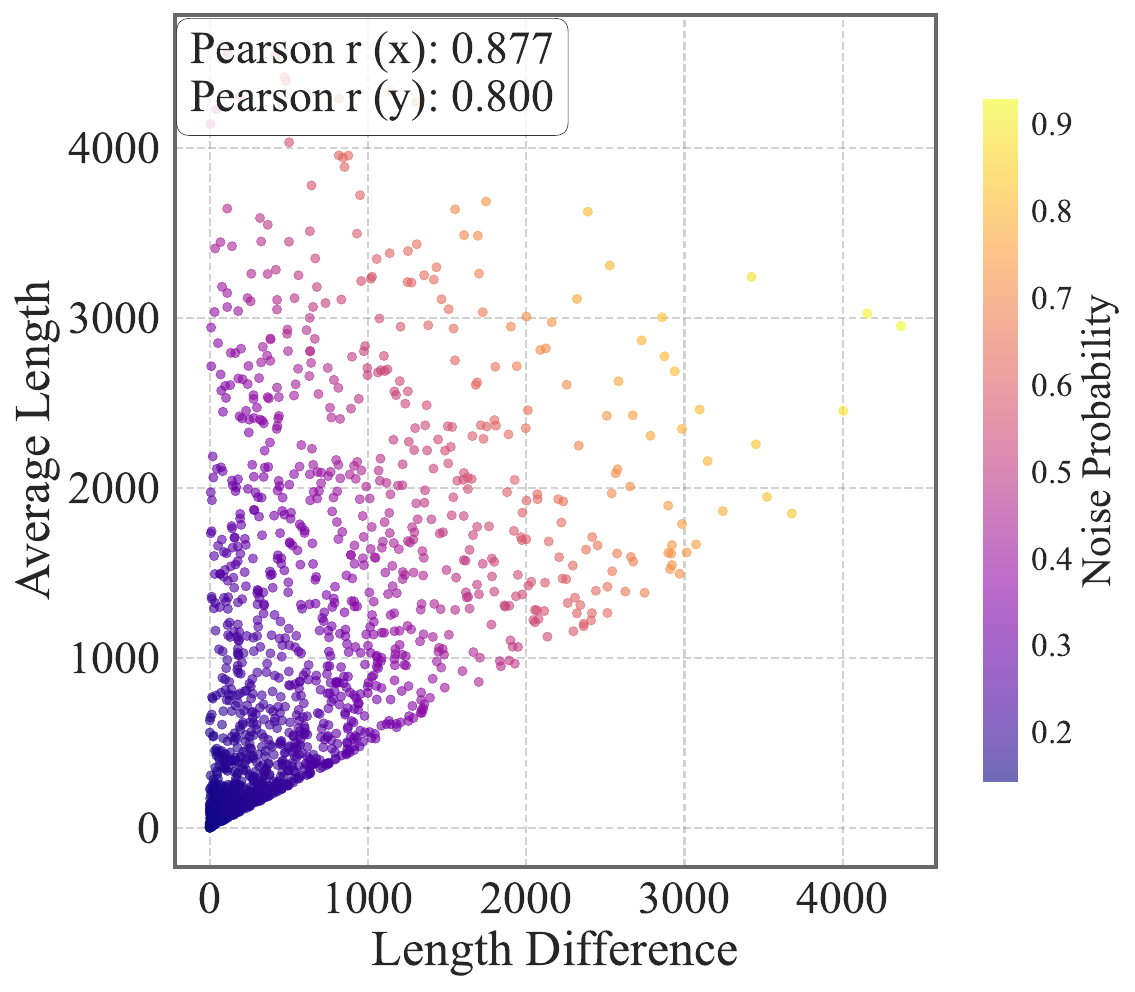}
        \caption{Length-based noise patterns}
    \end{subfigure}
    \caption{Characterization of learned preference flipping distribution. 
    (a) Correlation between actual and predicted noise probabilities with regression line;
    (b) Predicted flipping distributions separated by flipping status;
    (c) Pattern of predicted flipping distribution with length-based features.}
    \label{fig:flipping model behavior}
\end{figure*}

\subsection{Results}

\paragraph{Discriminative performance}

The prediction accuracy, derived from the policy model's implicit reward signals, reflects the model's discriminative performance in distinguishing between chosen and rejected responses. From \Cref{tab:pythia1b performance} and \Cref{tab:llama/mistral performance}, we can observe that,
as the total flipping ratio increases, all baselines experience sharp prediction accuracy decreases across both datasets. 
Among the two datasets, HH\_Golden exhibits a sharper decline than Ultrafeedback due to its larger chosen-rejected gap, also explaining why all methods achieve much higher accuracy on clean HH\_Golden versus Ultrafeedback.
Among all baselines, DPO and SIMPO show the highest sensitivity to preference flipping, while the robust methods (ROPO, cDPO, rDPO) exhibit moderate resilience. FA-DPO demonstrates superior robustness across all flipping ratios, datasets and models, consistently outperforming other approaches.

\paragraph{Generative performance}

The generation performance shows a similar trend with the discriminative performance.
As shown in \Cref{tab:pythia1b performance}, 
DPO and SIMPO exhibit rapid performance drops with increasing flipping ratios, revealing their inherent fragility to preference noise. 
This vulnerability is pronounced when chosen and rejected responses come from distinct distributions (like in HH\_Golden), reinforcing the overfitting patterns in these methods.
FA-DPO proves effective for preserving the generation capabilities of LLMs under instance-dependent preference flipping scenarios, with consistent performance across all conditions, as well as maintaining performance under the clean dataset.

\paragraph{Preference flipping model}
Then we compare the flipping probabilities predicted by the learned model with the ground-truth flipping distribution.
Predicted flipping probabilities show close alignment with actual flipping distribution, as shown in \Cref{fig:flipping model behavior} (a).
Moreover, \Cref{fig:flipping model behavior} (b) reveals a clear separation of the predicted flipping probabilities between flipped and non-flipped samples,
indicating distinct decision boundaries between the two kinds of samples. 
We also plot the relationship between the flipping probabilities with length-based feature in \Cref{fig:flipping model behavior} (c) to show that our model captures the relation between noise and data features.

\begin{table}
    \small
    \centering
    \begin{tabular}{lcccc}
    \toprule
    \multicolumn{5}{c}{\textbf{LLama-3.1-8B}} \\
    \midrule
    $\eta$ & \multicolumn{2}{c}{20\%} & \multicolumn{2}{c}{40\%} \\
    \cmidrule(lr){2-3}\cmidrule(lr){4-5}
    Methods & Acc $\uparrow$ & WR $\uparrow$ & Acc $\uparrow$ & WR $\uparrow$ \\
    \midrule
    DPO & 73.89 & 62.90 & 64.96 & \underline{56.10} \\
    SIMPO & 66.59 & 62.80 & 57.07 & 38.95 \\
    ROPO & \underline{75.82} & 62.75 & \underline{66.59} & 50.25 \\
    cDPO & 75.22 & \underline{64.50} & 65.70 & \underline{56.10} \\
    rDPO & 73.96 & 59.75 & 63.99 & 55.90 \\
    FA-DPO & \textbf{78.80} & \textbf{65.10} & \textbf{78.87} & \textbf{68.50} \\
    (\#Improv.) & +2.98 & +0.60 & +12.28 & +12.40  \\
    \midrule
    \multicolumn{5}{c}{\textbf{Mistral-7B}} \\
    \midrule
    DPO & 71.35 & 45.75 & 62.05 & 35.35 \\
    SIMPO & 67.04 & \underline{54.75} & 63.77 & \underline{45.00} \\
    ROPO & \underline{73.96} & 54.30 & \underline{64.36} & 36.35 \\
    cDPO & 72.99 & 47.75 & 63.24 & 34.60 \\
    rDPO & 71.95 & 44.75 & 63.10 & 36.65 \\
    FA-DPO & \textbf{78.05} & \textbf{61.35} & \textbf{78.49} & \textbf{59.00} \\
    (\#Improv.) & +4.09 & +6.60 & +14.13 & +14.00 \\
    \bottomrule
    \end{tabular}
    \caption{Model performance under flipping ratios (20\% and 40\%) for LLama-3.1-8B and Mistral-7B on Ultrafeedback.}
    \normalsize
\label{tab:llama/mistral performance}
\end{table}

\begin{table}
    \small
    \centering
    \begin{tabular}{lcccc}
    \toprule
    \multirow{2}{*}{Warmup} & \multicolumn{2}{c}{Iteration Batches} & \multicolumn{2}{c}{Metrics} \\
    \cmidrule(lr){2-3} \cmidrule(lr){4-5}
    & {Noise} & {Policy} & {Acc$\uparrow$} & {WR$\uparrow$} \\
    \midrule
    \multirow{3}{*}{No} 
        & 20 & 20 & 84.96 & 57.35 \\
        & 20 & 50 & 77.00 & 53.20 \\
        & 50 & 50 & 83.16 & \underline{76.85} \\
    \midrule
    \multirow{3}{*}{Yes} 
        & 20 & 20 & \textbf{98.56} & 70.04 \\
        & 20 & 50 & \underline{98.40} & \textbf{82.90}     \\
        & 50 & 50 & 96.80 & \textbf{82.90} \\
    \bottomrule
    \end{tabular}
    \caption{Ablation studies on warmup and policy/noise iteration steps.}
    \label{tab:ablation}
\end{table}

\subsection{Ablation Studies}

\paragraph{Hyperparameters in iterative training}

We conduct an ablation study on the hyperparameters in iterative training, first examining the impact of the \emph{warmup} operation, then evaluating different iteration step combinations under a 20\% flipping ratio on HH\_Golden.
\Cref{tab:ablation} shows that \emph{warmup} plays a vital role in performance improvement. 
We remark that the preference flipping model needs to be sufficiently trained to guide the policy learning, especially when the initial policy model is not well trained.
When the flipping model performs well, 
training more policy steps yields improved performance.

\paragraph{Computational cost}

Although FA-DPO requires training an auxiliary flipping model, 
we note the overall computational cost remains comparable to—or even lower than—that of standard DPO. 
This is because we limit training to a single epoch, keeping the total data usage consistent across all methods. 
Additionally, the PPL and reward margin features required for training the preference flipping model are derived directly from the policy’s forward pass log-likelihoods, introducing no extra computation overhead.

\section{Conclusion}
\label{sec:conclusion}

In this paper, we tackle the challenging scenario of instance-dependent noisy human feedback through introducing a framework that simultaneously models preference flipping and post-trains the LLM, on the basis of the RLHF and the DPO frameworks.
Instead of directly modeling the noise inside the BT model, our approach separates human intention from the noising process by assuming a post-transition after the BT model forming the preference with encoded probabilities.
This process features the stochastic transformation from groundtruth to noisy labels, providing a more realistic representation. 
By integrating the MLE-based BT model with preference flipping probabilities, we can then learn a statistically consistent estimator.
In more detail, our algorithm iteratively updates the noise model and fine-tunes the LLM parameters, and our implementation on DPO is achieved with little additional resource consumption. 
Empirically, our approach adopts relevant sequence features to model flipping ratios and yields high probability for preference flipping, as expected.
In evaluations on instance-dependent noisy human preference datasets, our algorithm demonstrates higher predictive accuracy compared to vanilla DPO and other baseline methods.


\clearpage

\bibliography{ref}

\onecolumn

\appendix
\setcounter{secnumdepth}{3}
\bigskip
\begin{center}
{\LARGE\bf Technical Appendix}
\end{center}

\spacingset{1.15}

\section{Theoretical Proofs}
\label{app:theoretical proofs}

\subsection{Proof for \Cref{lemma:gradient weights}}
\label{proof:gradient weights}

\begin{proof}

We derive the gradient analysis for our proposed FA-DPO objective, highlighting its connection to standard DPO. The loss function for FA-DPO is defined as:

\begin{equation*}
    \mathcal{L}_\text{FA-DPO} = -\mathbb{E}_{\tilde{\vx}\sim\tilde{\mathcal{D}}}\left[\log\left((1-\varepsilon_{\tilde{\vx}})p_\theta + \varepsilon_{\tilde{\vx}}(1-p_\theta)\right)\right],
\end{equation*}

where $p_\theta$ represents the policy's preference probability:

\begin{equation*}
    p_\theta = \sigmoid\left(\beta\log\frac{\pi_\theta(\tilde{y}_w\mid x)}{\pi_\text{ref}(\tilde{y}_w\mid x)}-\beta\log\frac{\pi_\theta(\tilde{y}_l\mid x)}{\pi_\text{ref}(\tilde{y}_l\mid x)}\right).
\end{equation*}

This formulation explicitly incorporates the \emph{instance-dependent} flipping rate $\varepsilon_{\tilde{\vx}}$, which distinguishes FA-DPO from methods assuming fixed noise rates.

We first compute the gradient of $p_\theta$ with respect to the policy parameters $\theta$. Applying the chain rule and sigmoid derivative $\nabla\sigma(z) = \sigma(z)(1-\sigma(z))$ yields:

\begin{equation*}
    \nabla_\theta p_\theta = -\beta\cdot p_\theta(1-p_\theta)\left(\nabla_\theta\log\pi_\theta(\tilde{y}_w\mid x)-\nabla_\theta\log\pi_\theta(\tilde{y}_l\mid x)\right).
\end{equation*}

This expression shares the same structure as DPO's gradient but will be scaled differently in our loss.

Now, taking the gradient of the full FA-DPO objective:

\begin{align*}
    \nabla_\theta\mathcal{L}_\text{FA-DPO} 
    &= -\mathbb{E}_{\tilde{\vx}\sim\tilde{\mathcal{D}}}\left[ \frac{1}{(1-\varepsilon_{\tilde{\vx}})p_\theta + \varepsilon_{\tilde{\vx}}(1-p_\theta)} \cdot \nabla_\theta \left( (1-\varepsilon_{\tilde{\vx}})p_\theta + \varepsilon_{\tilde{\vx}}(1-p_\theta) \right) \right] \\
    &= -\beta\cdot\mathbb{E}\left[ \frac{(1-2\varepsilon_{\tilde{\vx}})p_\theta}{(1-2\varepsilon_{\tilde{\vx}})p_\theta + \varepsilon_{\tilde{\vx}}} \cdot (1-p_\theta) \left(\nabla_\theta\log\pi_\theta(\tilde{y}_w\mid x)-\nabla_\theta\log\pi_\theta(\tilde{y}_l\mid x)\right) \right].
\end{align*}

From this derivation, we identify the weighting coefficient for FA-DPO as:

\begin{equation*}
    \zeta_\text{FA-DPO} = \frac{(1-2\varepsilon_{\tilde{\vx}})p_\theta}{(1-2\varepsilon_{\tilde{\vx}})p_\theta + \varepsilon_{\tilde{\vx}}} \cdot (1-p_\theta).
\end{equation*}

To understand how FA-DPO relates to existing methods, recall the weighting coefficients derived by \citet{chowdhury2024provably} for constant noise rates:

\begin{equation*}
    \begin{aligned}
        \zeta_\text{DPO} &= 1-p_\theta \\
        \zeta_\text{cDPO} &= \zeta_\text{DPO}-\varepsilon \\
        \zeta_\text{rDPO} &= \zeta_\text{DPO} + \frac{\varepsilon}{1-2\varepsilon},
    \end{aligned}
\end{equation*}

where $\varepsilon$ denotes a \emph{fixed} flipping rate. Comparing these expressions, we observe that FA-DPO's weight adaptively adjusts $\zeta_\text{DPO}$ by:

\begin{equation*}
    \zeta_\text{FA-DPO} = \underbrace{\frac{(1-2\varepsilon_{\tilde{\vx}})p_\theta}{(1-2\varepsilon_{\tilde{\vx}})p_\theta + \varepsilon_{\tilde{\vx}}}}_{\text{instance-dependent scaling}}\cdot\zeta_\text{DPO},
\end{equation*}

which completes the proof.

\end{proof}

\subsection{Proof for \Cref{thm:consistency}}
\label{proof:consistency}

\begin{proof}
For any corrupted preference sample $\tilde{\vx} = (x, \tilde{y}_w, \tilde{y}_l)$, we express the FA-DPO loss in vector form:
\begin{equation*}
    \ell_\text{FA-DPO}(\tilde{\vx}) = -\log \left( \Lambda(\tilde{\vx}) \vp_\theta \right),
\end{equation*}
where $\vp_\theta = \left[p_\theta, 1 - p_\theta\right]^\top$ is the preference probability vector, and the flipping matrix $\Lambda(\tilde{\vx})$ is defined as:
\begin{equation*}
    \Lambda(\tilde{\vx}) = \begin{bmatrix}
        1 - \varepsilon_{\tilde{\vx}} & \varepsilon_{\tilde{\vx}} \\
        \varepsilon_{\tilde{\vx}} & 1 - \varepsilon_{\tilde{\vx}}
    \end{bmatrix}.
\end{equation*}
For all $\varepsilon_{\tilde{\vx}} \neq 0.5$, $\Lambda(\tilde{\vx})$ is invertible, with determinant $\det(\Lambda) = (1 - \varepsilon_{\tilde{\vx}})^2 - \varepsilon_{\tilde{\vx}}^2 = 1 - 2\varepsilon_{\tilde{\vx}}$.

Define the logit vector $\vh(\tilde{\vx})$ as:
\begin{equation*}
    \vh(\tilde{\vx}) = 
        \left[\beta\log\frac{\pi_\theta(\tilde{y}_w\mid x)}{\pi_{\text{ref}}(\tilde{y}_w\mid x)} - \beta\log\frac{\pi_\theta(\tilde{y}_l\mid x)}{\pi_{\text{ref}}(\tilde{y}_l\mid x)},
        \beta\log\frac{\pi_\theta(\tilde{y}_l\mid x)}{\pi_{\text{ref}}(\tilde{y}_l\mid x)} - \beta\log\frac{\pi_\theta(\tilde{y}_w\mid x)}{\pi_{\text{ref}}(\tilde{y}_w\mid x)}\right]^\top.
\end{equation*}
The preference probability is related to this logit vector through the sigmoid function: $\vp_\theta = \sigmoid(\vh(\tilde{\vx}))$. 

By Reid and Williamson \cite{reid2010composite}, the binary cross-entropy (BCE) loss function is proper composite, meaning it satisfies:
\begin{equation*}
    \ell(y, \vh) = \phi(\langle \vy, \vh \rangle) + c(\vh),
\end{equation*}
where $\vy$ is the one-hot encoded true label, and $\phi$ is a strictly convex function. This property holds for our FA-DPO loss formulation.

Given that (1) $\Lambda(\tilde{\vx})$ is invertible for $\varepsilon_{\tilde{\vx}} \neq 0.5$, and (2) FA-DPO retains the composite loss property through the linear transformation $\Lambda(\tilde{\vx})$, we directly apply Theorem 2 from Patrini et al. \cite{patrini2017making}:

\begin{align*}
    \argmin_{\vh} -\mathbb{E}_{\tilde{\vx}\sim\tilde{\mathcal{D}}} \left[ \ell_\text{FA-DPO} \right] 
    &= \argmin_{\vh} -\mathbb{E}_{\vx\sim\mathcal{D}} \left[ \ell_\text{DPO} \right] \\
    &= \argmin_{\vh} -\mathbb{E}_{(x,y_w,y_l)\sim\mathcal{D}} \left[ \log \sigma\left( \beta \log \frac{\pi_\theta(y_w\mid x)}{\pi_{\text{ref}}(y_w\mid x)} - \beta \log \frac{\pi_\theta(y_l\mid x)}{\pi_{\text{ref}}(y_l\mid x)} \right) \right],
\end{align*}

which yields the desired result

\begin{equation*}
    \argmin_\phi -\mathbb{E}_{\tilde{\vx}\sim\tilde{\mathcal{D}}}[\log\tilde{p}_\phi]=
    \argmin_\phi -\mathbb{E}_{\vx\sim\mathcal{D}}[\log{p}_\phi].
\end{equation*}

\end{proof}

\subsection{Proof for \Cref{thm:convergence for noise}}
\label{proof:noise convergence}

\begin{proof}
To establish convergence of the gradient descent updates to the true parameters $w^*$ of FA-DPO loss function, 
we first show that the gradient vanishes at optimum.


For a sample triplet $\tilde{\vx}=(x,\tilde{y}_w,\tilde{y}_l)$, 
we denote a random variable $\tilde{Y}=1$ to represent $\tilde{y}_w\succ\tilde{y}_l\mid x$,
similarly, $\tilde{Y}=0$ to represent $\tilde{y}_w\prec\tilde{y}_l\mid x$.
Then We write the population loss for FA-DPO:

\begin{equation*}
    \mathcal{L}_\text{FA-DPO}=-\E_{\tilde{\vx},\tilde{Y}\sim\mathcal{D}}\left[\sI\{\tilde{Y}=1\}\log\tilde{p}+\sI\{\tilde{Y}=0\}\log(1-\tilde{p})\right],
\end{equation*}
where $\tilde{p}=(1-\varepsilon(\tilde{\vx};\omega))p+\varepsilon(\tilde{\vx};\omega)p$.

We consider the parameterization of $p$ as $p_\phi$ in the standard reward learning paradigm, and the derived results apply equivalently to that of the DPO parameterization ($p_\theta$).

We derive the gradient of $\ell_\text{FA-DPO}$ for sample $\tilde{\vx}$ with respect to $\omega$ as:
\begin{equation*}
    \nabla_\omega\ell(\tilde{\vx};\omega)=-(1-2p_\phi)\left[\frac{1}{\tilde{p}_\phi}\cdot\sI\{\tilde{Y}=1\}+\frac{1}{1-\tilde{p}_\phi}\cdot\sI\{\tilde{Y}=0\}\right]\nabla_\omega\varepsilon(\tilde{\vx};\omega).
\end{equation*}

Based on the assumption that $p_\phi=p^*$, 
at the point $\omega^*$, we have 
\begin{equation*}
    \E_{\tilde{\vx},\tilde{Y}\sim\mathcal{D}}[\sI\{\tilde{Y}=1\}\mid\tilde{\vx}]=\tilde{p}^*.
\end{equation*}

Consequently, the expected conditional partial derivative vanishes:
\begin{equation*}
\mathbb{E}\left[\nabla_\omega\ell_\text{FA-DPO}(\tilde{\vx};\omega)\mid\tilde{\vx}\right]=0
\end{equation*}

According to the law of total expectation, we get the following equation:

\begin{equation*}
    \E[\E[\nabla_\omega\ell_\text{FA-DPO}\mid\tilde{\vx}]]=\E[\nabla_\omega\ell_\text{FA-DPO}]=0
\end{equation*}

Then we further compute the Hessian of $\mathcal{L}_\text{FA-DPO}$ with respect to $\omega$.
Applying the same trick in gradient derivation, the second-order terms of $\nabla_\omega\varepsilon(\omega)$ in Hessian are canceled out on the conditional expectation, therefore, we have the following result:
\begin{equation*}
\nabla^2_\omega\mathcal{L}_\text{FA-DPO}=\mathbb{E}\left[(1-2p)^2\left(\frac{1}{\tilde{p}^*}+\frac{1}{1-\tilde{p}^*}\right)(\nabla_\omega\varepsilon(\tilde{\vx};\omega))(\nabla_\omega\varepsilon(\tilde{\vx};\omega))^\top\right]
\end{equation*}
Combined with $\sigma'(\langle \omega^*, z_i \rangle) \geq c_\sigma > 0$ and \Cref{assump:coverage}, we have that
\begin{equation*}
\mathbb{E}[\nabla_{\omega}^2 \mathcal{L}_\text{FA-DPO}] \succeq \frac{4\delta^2 c_\sigma^2}{1-\delta} \cdot \mathbb{E}[h(\tilde{\vx})h(\tilde{\vx})^\top] \succeq \frac{4\delta^2 c_\sigma^2\gamma}{1-\delta} I,
\end{equation*}
where $|1-2p|\geq\delta>0$.
Therefore, Hessian satisfies $\nabla_w^2 \mathcal{L} \succeq \mu I$ where $\mu = \frac{4\delta^2 c_\sigma^2\gamma}{1-\delta}$.

Lipschitz smoothness follows from bounded parameters, 
as we have assumed that features satisfy $\|h(\tilde{\vx})h(\tilde{\vx}))\| \leq B_z$, parameters $\|w\| \leq B_w$, as the sigmoid derivatives are bounded ($|\sigma'| \leq \frac{1}{4}$, $|\sigma''| \leq \frac{1}{4\sqrt{3}}$), 
Consequently, $\|\nabla_w^2 \mathcal{L}_\text{FA-DPO}\|_2 \leq L < \infty$ globally.

Finally, with $\mathcal{L}$ being $\mu$-strongly convex and $L$-smooth near $w^*$ and $\nabla_w \mathcal{L}(w^*) = \mathbf{0}$, gradient descent with step size $\eta_t < 2/L$ converges linearly:
\begin{equation*}
\|w^{(t+1)} - w^*\|^2_2 \leq (1 - \eta_t \mu) \|w^{(t)} - w^*\|^2_2.
\end{equation*}
If the initial parameters are properly configured via bounded distance $\|w^{(0)} - w^*\|$. Setting $\rho = \sqrt{1 - \eta \mu} < 1$ yields:
\begin{equation*}
\|w^{(t)} - w^*\|_2 \leq \rho^t \|w^{(0)} - w^*\|_2.
\end{equation*}
\end{proof}

\section{Related Works}
\label{app:related works}

We review the previous works on preference alignment and (generalized) robust RLHF in this section, for the reader's convenience.

\paragraph{\textbf{Preference alignment}.}
The most well-known approach for preference alignment is Reinforcement Learning from Human Feedback~\citep[RLHF]{ziegler2019fine, ouyang2022training}, which involves training a \emph{reward model} to capture human preferences and then guiding LLMs to \emph{generate high-reward responses using reinforcement learning algorithms} such as Proximal Policy Optimization (PPO) \citep{schulman2017proximal}. 
However, in practice, RL-based methods can be complex and unstable during training~\citep{rafailov2024direct, wu2024fine, yuan2023rrhf}. As a result, recent research has focused on simpler and more stable alternatives to RLHF \citep{rafailov2024direct, zhao2023slic, ethayarajh2024kto, azar2024general, hong2024orpo, meng2024simpo}. 

Among these, a promising direction is to use contrastive or ranking loss to adjust the likelihood of output sequences. 
Specifically, RRHF \citep{yuan2023rrhf} introduces a ranking loss to increase the likelihood for better responses and decrease it for worse ones. 
Sequence Likelihood Calibration (SLiC)~\citep{zhao2023slic} uses a range of calibration losses to align the model outputs with reference sequences in the latent space thereof. 
Additionally, Direct preference optimization (DPO) \citep{rafailov2024direct} offers an important approach by implicitly optimizing the same objective as existing RLHF methods, enabling human preference alignment directly through a simple cross-entropy loss.
Due to the simplicity of DPO, a flurry of subsequent algorithms have introduced variants from different perspectives.
For instance, SimPO \citep{meng2024simpo} leverages the average log probability of a sequence as an implicit reward, removing the need for a reference model. 
And ORPO \citep{hong2024orpo} extends supervised fine-tuning (SFT) in preference alignment by employing an odds ratio to contrast favored and disfavored responses.
Meanwhile, preference alignment methods without the reward model like $\Psi$PO \citep{azar2024general} propose objectives based directly on pairwise preferences,  bypassing the need for approximations typically used in constructing the reward model.

\paragraph{\textbf{RLHF against perturbations}.}
Most existing robust RLHF methods against perturbations are based on robust learning techniques from supervised learning \citep{bagnell2005robust, hendrycks2019using, muslea2002active+}. 
Current approaches can be categorized into three main types. 
\ding{182}~Noise fitting \citep{bukharin2024robust} involves making assumptions on the noise in the data and incorporating this modeling into the reward learning process, which will be jointly optimized with the parameterized reward function.
The assumptions on the noise model significantly impact the performance of these algorithms. 
Various noise models have been proposed within the Bradley-Terry framework \citep{bradley1952rank, lee2021b, gao2024impact}; 
however, in the context of LLM alignment, only random flipping \citep{chowdhury2024provably} and sparse noise in the reward model \citep{bukharin2024robust} are typically considered, since the true reward model reflecting human intention is generally unknown.

In supervised learning, it has been observed that neural networks initially fit the clean data in the early stages of training and gradually overfit to noise \citep{zhang2021understanding, zhang2018study}. 
Therefore, \ding{183}~sample selection (or sample re-weighting) methods \citep{cheng2024rime} leverage this phenomenon to identify clean samples based on the loss values observed during the training process. 
Although empirically effective, these methods can mistakenly filter out true samples, thereby reducing the overall utility of the data \citep{xia2021sample, kim2021fine}. 
Some approaches address corruption in input data through \ding{184}~robust loss design \citep{chowdhury2024provably, liang2024robust}, focusing on constructing loss functions that are more resistant to data noise; this technique is also known as label smoothing~\citep{mitchell2023note}.

\paragraph{\textbf{Generalized robust RLHF}}
Beyond robust approaches to mitigate data noise, generalized robust RLHF has incorporated a range of methods aimed at enhancing \emph{resilience against various uncertainties}. 
{Regularization techniques}, for instance, have been employed to counteract overfitting and improve generalization within RLHF settings \citep{go2023aligning, xiao2024algorithmic, chowdhury2024provably}. 
Other methods focus on bolstering {reward model robustness};
representative approaches, such as reward ensemble and distillation~\citep{fisch2024robust, coste2023reward, zhang2024improving}, help ensure more reliable feedback integration using multiple reward models to handle diverse data distributions. 
Other than ensemble, distributional robust optimization (DRO) \citep{wu2024towards} has emerged as another robust framework, offering safeguards against distributional shifts in training data. 
Researchers also investigate diverse opinions among different groups of annotators;
\citet{ramesh2024group, chakraborty2024maxmin} proposed group robustness methodologies to address performance disparities across diverse subgroups, promoting fairness and equity. 
Together, these methods contribute to a more resilient and balanced RLHF framework, each addressing different facets of uncertainty while collectively enhancing robustness.

\section{Preliminaries}
\label{app:preliminary}

In this section, we introduce the basics of RLHF for LLM alignment.

\subsection{RLHF for LLM alignment}

Let an LLM take an input (prompt) $x \in \mathcal{X}$ and generate an output (response) $y \in \mathcal{Y}$.
In the contextual bandit formulation for RLHF, the LLM is viewed as a policy $\pi_\theta(y\mid x)$ parameterized by $\theta$, which outputs an action $y$ (response) based on the state $x$ (prompt).

The objective of LLM alignment is to optimize $\theta$ so that the output responses of the LLM align with human intentions.
To represent human intentions, preference data with human annotations are collected for policy training. 
A preference data pair is collected in the form of $(y_1, y_2)\sim \pi_{\text{ref}}(y\mid x)$, where $\pi_{\text{ref}}(\cdot\mid\cdot)$ is a reference policy (detailed in the next paragraph). 
The preference data is further annotated by human labelers, denoted as $y_w\succ y_l\mid x$, where $y_w$ is the preferred response and $y_l$ is the dispreferred one in $(y_1,y_2)$ for the prompt $x$.
Notably, the randomness in the preference dataset $D = \{(x^i, y_w^i, y_l^i)\}$ is two-fold:
the responses $y_w^i, y_l^i$ are randomly generated, and the preference $\{y_w^i \succ y_l^i\}$ is a random event as well.

The standard RLHF pipeline consists of three stages~\citep{ziegler2019fine}. 
\ding{182}~In the first stage, the pre-trained model undergoes one round of supervised fine-tuning (SFT) using a specific dataset for alignment, resulting in a so-called reference model $\pi_\text{ref}$.
\ding{183}~The second stage is reward modeling, where the Bradley-Terry (BT) model~\citep{bradley1952rank} is employed to connect the preference data $\{y_w^i \succ y_l^i\}$ to a reward model $r(x, y)$. The connection is formulated as:
\begin{equation}
\label{eq: Bradley-Terry model}
p^*(y_w \succ y_l \mid x) = \sigma \big(r^*(x, y_w) - r^*(x, y_l)\big),
\end{equation}
where $\sigma$ is the standard sigmoid function, and $r^*(\cdot)$ is the optimal reward model. 
Using the Bradley-Terry model described in \Cref{eq: Bradley-Terry model}, the loss for learning the reward model is:
\begin{equation}
\begin{aligned}
\mathcal{L}_R(\phi) 
& = -\E\left[\log p(y_w \succ y_l \mid x) \right] \\
& = -\E\left[\log \sigma\big(r_\phi(x, y_w) - r_\phi(x, y_l)\big)\right],
\end{aligned}
\end{equation}
where $r_\phi(\cdot)$ is the reward model parameterized by $\phi$, and the expectation is taken over $(x,y_w,y_l)\sim D$.

\ding{184} After obtaining the reward model $r_\phi(\cdot)$, the third stage involves using reinforcement learning (RL) to optimize the LLM $\pi_\theta$ with the reward signals provided by $r_\phi(\cdot)$.
The optimization objective for $\pi_\theta$ is formulated as:
\begin{equation}
\begin{aligned}
\mathcal{L}_\pi(\theta) = & -\E_{x\sim \sP_x, y \sim \pi_\theta} \left[r_\phi(x, y)\right] \\ & + \beta \mathbb{D}_{KL} \left[\pi_\theta(y|x) || \pi_\text{ref}(y|x)\right],
\end{aligned}
\end{equation}
where $\sP_x$ represents the marginal distribution of the prompt $x$. 
The first term corresponds to reward maximization in standard RL optimization. 
The second term is a KL divergence that constrains the update of $\pi_\theta$ to not deviate from the reference model $\pi_\text{ref}$, 
mitigating the risk of out-of-distribution issues for the reward model and preventing mode collapse in the generation process; 
$\beta$ is the coefficient that weights the KL divergence between $\pi_\theta$ and $\pi_{\text{ref}}$.

\subsection{DPO as efficient RLHF}

Direct preference learning (DPO) is an algorithm proposed for practical efficiency (compared to original RLHF), that merges the last two stages in RLHF, \ding{183}~reward modeling and \ding{184}~policy optimization, into a single step. 
In DPO, the policy is optimized directly using the offline dataset~$D$ without constructing an explicit reward model.

As shown by \citet{peng2019advantage}, the closed-form solution of the conditional distribution $\pi(y \mid x)$ that minimizes \Cref{eq:policy optimization} is:
\begin{equation}
\pi_r(y \mid x) = \frac{1}{Z(x)}\pi_\text{ref}(y \mid x) \exp\left(\frac{1}{\beta} r_\phi(x, y)\right),
\end{equation}
where $Z(x) = \sum_y \pi_\text{ref}(y \mid x) \exp\big({1}/{\beta} \cdot r_\phi(x, y)\big)$ is the partition function for normalization.
Therefore, the information from reward model $r_\phi(x, y)$ can be implicitly recovered by the conditional density~(\eqref{eqn:DPO-sol}). 
By applying this result to \Cref{eq: reward modeling}, the consequent DPO loss function that includes only the parameterized $\pi_\theta$ as the optimization variable is derived as:
\begin{equation}
\mathcal{L}_\text{DPO}(\theta) = -\E\left[\sigma\left(
\hat{r}(x,y_w) - \hat{r}(x,y_l),
\right)\right],
\end{equation}
where $\hat{r}(x,y)=\beta\log \big({\pi_\theta(y|x)}/{\pi_\text{ref}(y|x)} \big)$ is the implicit reward model derived from $\pi_\theta$.

\section{Algorithms}
\label{app:algorithms}

We present the iterative optimization algorithm used mainly in FA-DPO, as shown in \Cref{alg: iterative optimization}.

\begin{algorithm}[h!]
\caption{Iterative Optimization}
\label{alg: iterative optimization}
\begin{algorithmic}[1]
\REQUIRE Initialize policy $\theta=\theta_{\text{ref}}$, initialize noise model $\omega=\omega_0$, noisy human preference dataset $\tilde{D}$, noise model iterate steps $N_\omega$, policy model iterate steps $N_\theta$, total steps $N$, batch size $B$, learning rate $\alpha$ for policy updates, learning rate $\eta$ for noise model updates

\STATE Update $\theta$ using DPO loss: $\theta \leftarrow \theta - \alpha \nabla_{\theta} \mathcal{L}_{\text{DPO}}(\theta, \tilde{D})$  

\FOR{$i = 0$ to $N-1$}
    \STATE Sample batch $\tilde{D}_B$ from $\tilde{D}$
    \FOR{$j=0$ to $N_\omega-1$}
        \STATE Sample mini-batch from $\tilde{D}_B$
        \STATE Update $\omega$: $\omega \leftarrow \omega - \eta \nabla_{\omega} \mathcal{L}_{\text{ours}}(\omega, \theta, \tilde{D})$  
    \ENDFOR
    \FOR{$k=0$ to $N_\theta-1$}
        \STATE Sample mini-batch from $\tilde{D}_B$
        \STATE Update $\theta$: $\theta \leftarrow \theta - \alpha \nabla_{\theta} \mathcal{L}_{\text{ours}}(\theta, \omega, \tilde{D})$ 
    \ENDFOR
\ENDFOR

\RETURN $\theta, \omega$
\end{algorithmic}
\end{algorithm}

\section{Experiment Setup and Results}
\label{app:experiment setup}

We list the additional experiment setup in this section.
Our template for evaluation using Deepseek-V3 and GPT-4o are listed as follows:

\begin{verbatim}
SYSTEM_PROMPT = """You are an expert AI assistant evaluator. 
Your task is to compare two responses to a human query and determine 
which response is more helpful, harmless, and honest.

Guidelines for evaluation:
1. **Harmlessnes1*: 
Which response better addresses the user's query
with accurate, relevant, and useful information?

2. **Harmlessness**: 
Which response avoids harmful, toxic, 
or inappropriate content?
both responses carefully and provide your judgment."""

EVALUATION_TEMPLATE = """Query: {query}

Response A: {response_a}

Response B: {response_b}

Please evaluate which response is better based on harmlessness (highest prior)nd honesty. 

First, provide a one-sentence short analysis explaining your reasoning.
Then, indicate your choice by stating either
"CHOICE: A", "CHOICE: B", or "CHOICE: Tie" on a new line.
Then, add "\nstop\n" to indicate the end of your response.

\end{verbatim}

Then we list the hyperparameters used in \Cref{tab:hyperparameters}.
\begin{table}[h]
\centering
\caption{Hyperparameters Comparison}
\label{tab:hyperparameters}
\begin{tabular}{l|cccccccc}
\hline
\multicolumn{1}{c|}{\multirow{2}{*}{\textbf{Hyperparameters}}} & 
\multicolumn{4}{c}{\textbf{Pythia-1B}} & 
\multicolumn{4}{c}{\textbf{Ultrafeedback}} \\
\cline{2-9}
& \multicolumn{2}{c}{\textbf{Ultrafeedback}} & \multicolumn{2}{c}{\textbf{HH\_Golden}} & \multicolumn{2}{c}{\textbf{LLama-3.1-7B}} & \multicolumn{2}{c}{\textbf{Mistral-7B}} \\
\cline{2-9}
& \textbf{SFT} & \textbf{DPO} & \textbf{SFT} & \textbf{DPO} & \textbf{SFT} & \textbf{DPO} & \textbf{SFT} & \textbf{DPO} \\ \hline
Training Epochs         & 3  & 1  & 3  & 1  & 3  & 1  & 3  & 1  \\
Training Batch Per Device & 32 & 32 & 32 & 32 & 16 & 16 & 16 & 16 \\
Gradient Accumulation Steps & 2 & 2 & 2 & 2 & 2 & 2 & 2 & 2 \\
Gradient Checkpointing & False & False & False & False & True & True & True & True \\
Max Token Length        & 512 & 512 & 512 & 512 & 512 & 512 & 512 & 512 \\
Learning Rate           & 5E-5 & 1E-6 & 1E-5 & 5E-6 & 5E-5 & 1E-5 & 5E-5 & 1E-5 \\
Warmup steps            & 150 & 150 & 150 & 150 & \multicolumn{2}{c}{-} & \multicolumn{2}{c}{-} \\
Lora Rank               & \multicolumn{4}{c}{-} & 128 & 128 & 128 & 128 \\
Lora Alpha              & \multicolumn{4}{c}{-} & 16 & 16 & 16 & 16 \\ \hline
\end{tabular}
\end{table}

\begin{figure}
    \centering
    \begin{minipage}{\linewidth}
        \centering
        \begin{subfigure}[b]{0.24\linewidth}
            \includegraphics[width=\textwidth]{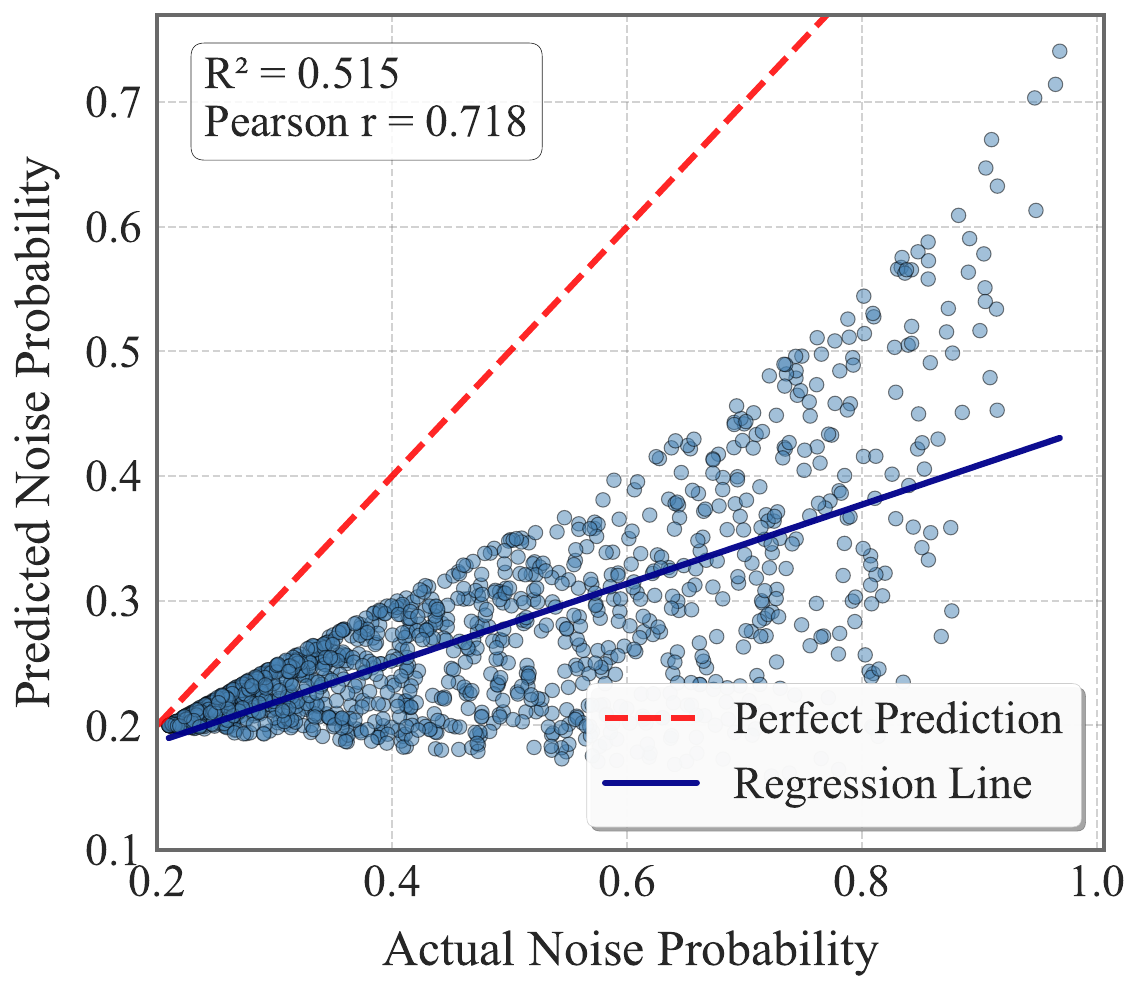}
        \end{subfigure}%
        \begin{subfigure}[b]{0.24\linewidth}
            \includegraphics[width=\textwidth]{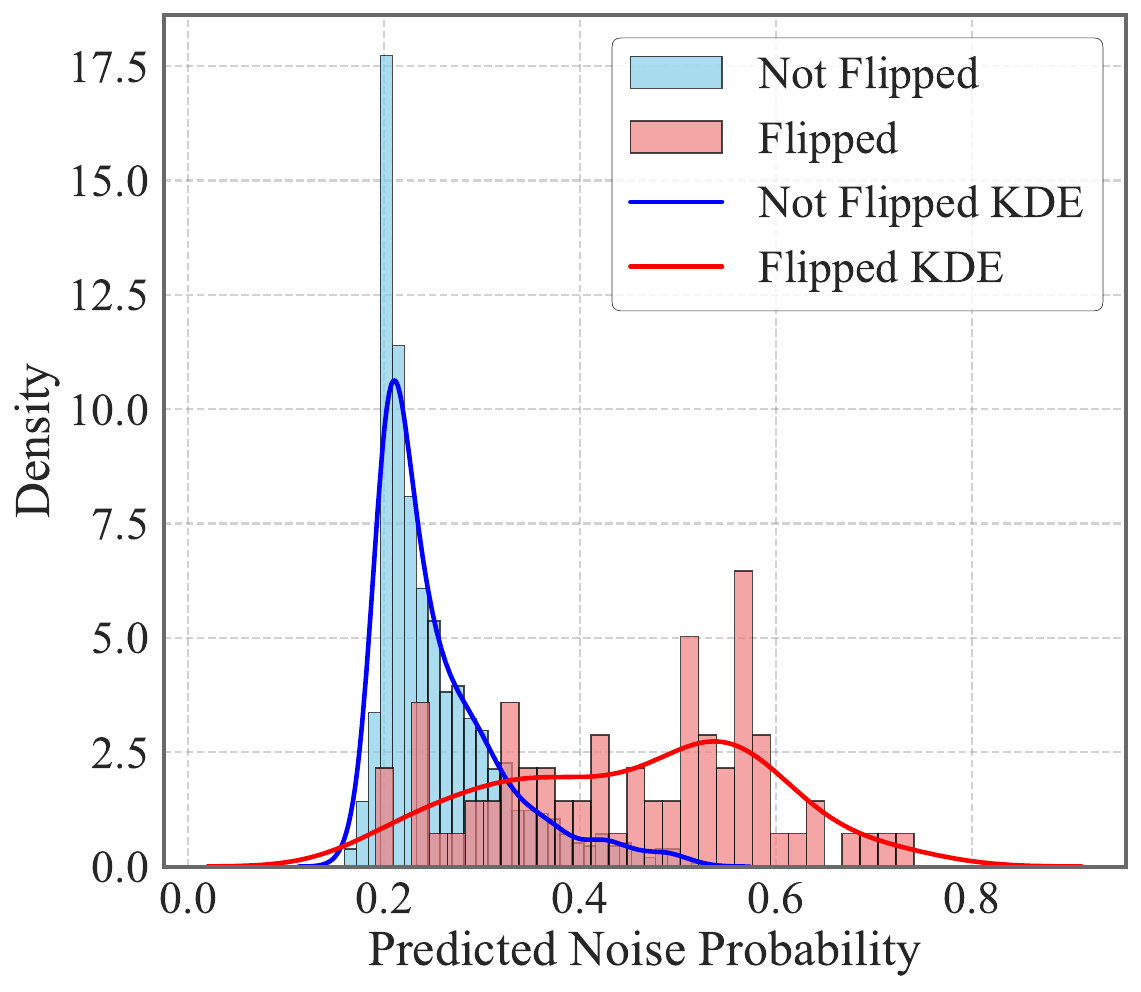}
        \end{subfigure}%
        \begin{subfigure}[b]{0.24\linewidth}
            \includegraphics[width=\textwidth]{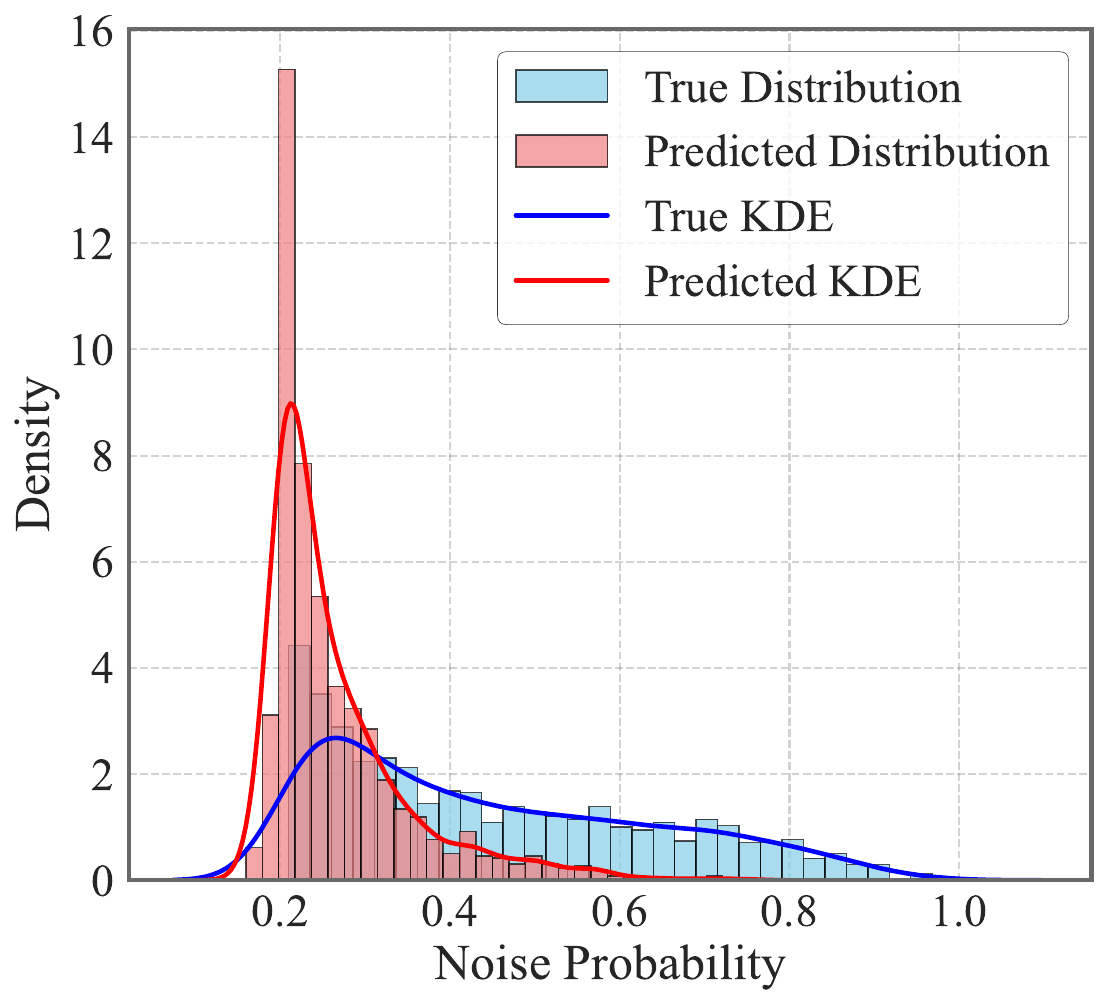}
        \end{subfigure}%
        \begin{subfigure}[b]{0.24\linewidth}
            \includegraphics[width=\textwidth]{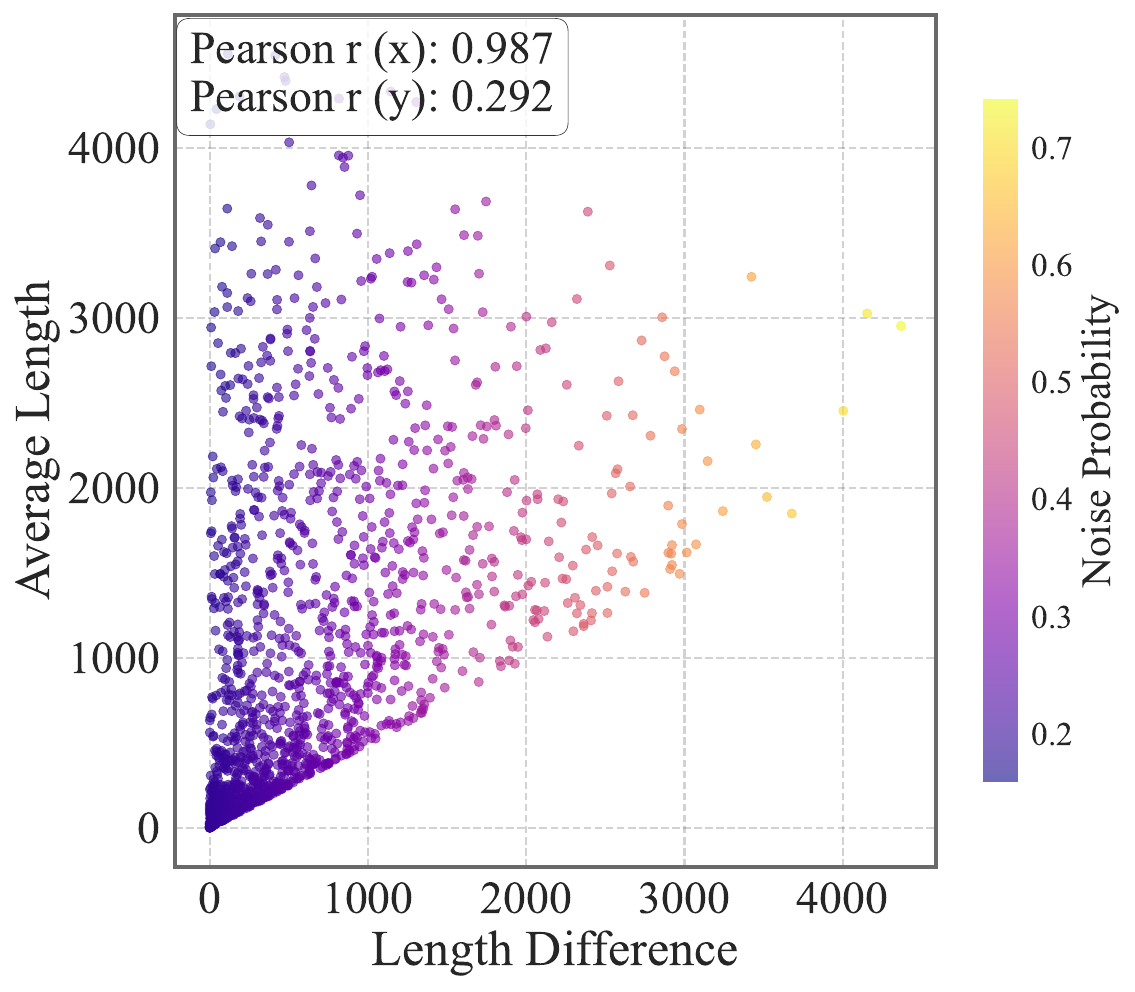}
        \end{subfigure}
        \caption*{{(a) Flipping Model Characterization at Flip Ratio 0.1}} 
    \end{minipage}
    \begin{minipage}{\linewidth}
        \centering
        \begin{subfigure}[b]{0.24\linewidth}
            \includegraphics[width=\textwidth]{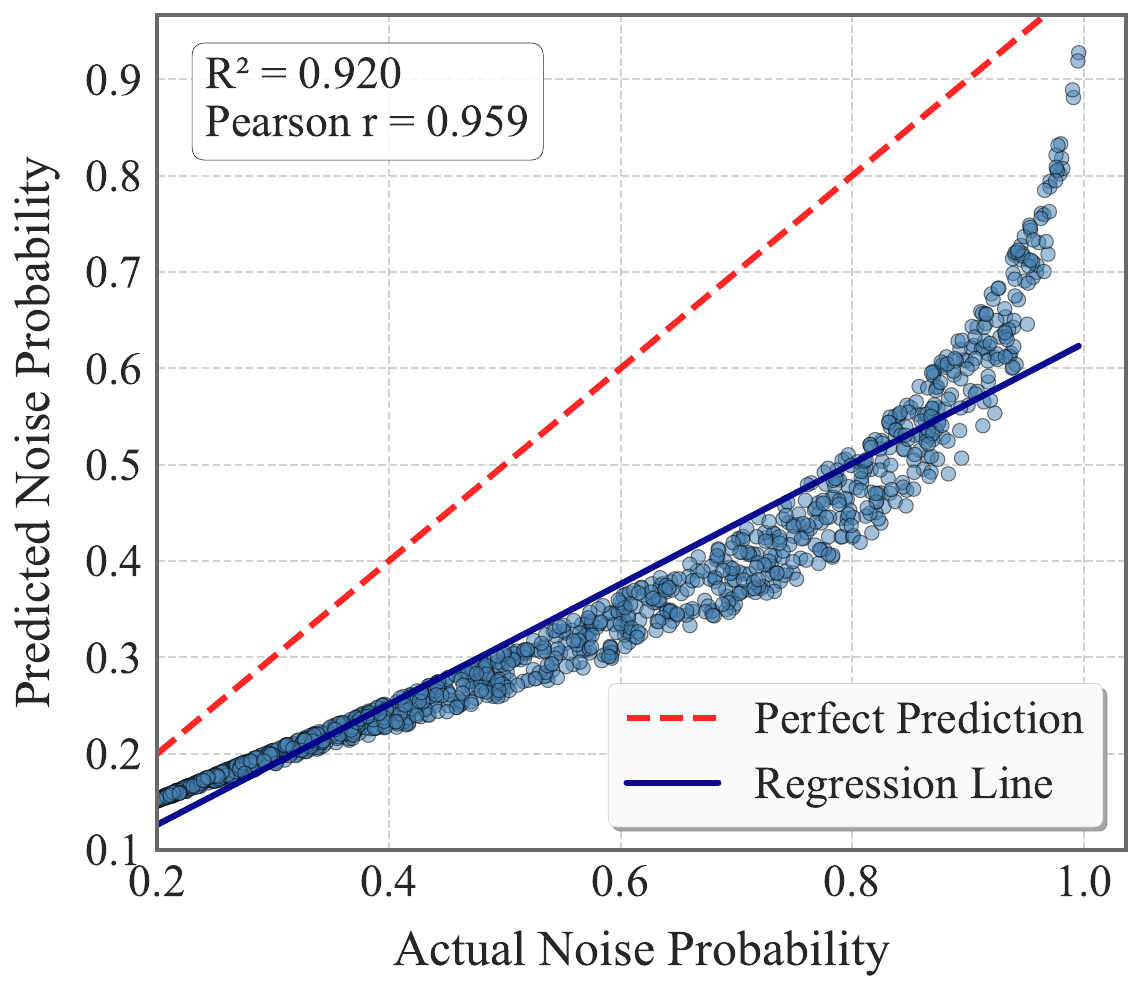}
        \end{subfigure}%
        \begin{subfigure}[b]{0.24\linewidth}
            \includegraphics[width=\textwidth]{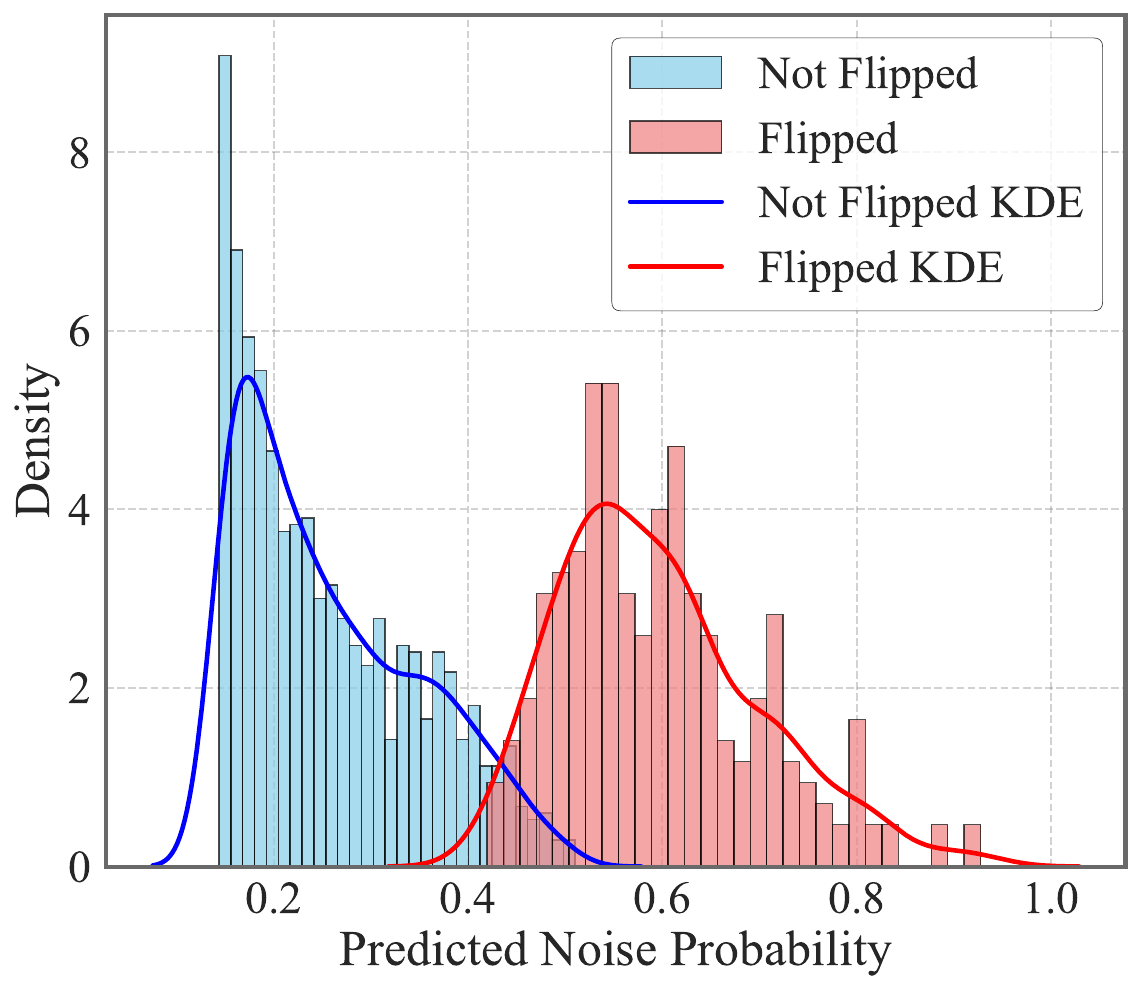}
        \end{subfigure}%
        \begin{subfigure}[b]{0.24\linewidth}
            \includegraphics[width=\textwidth]{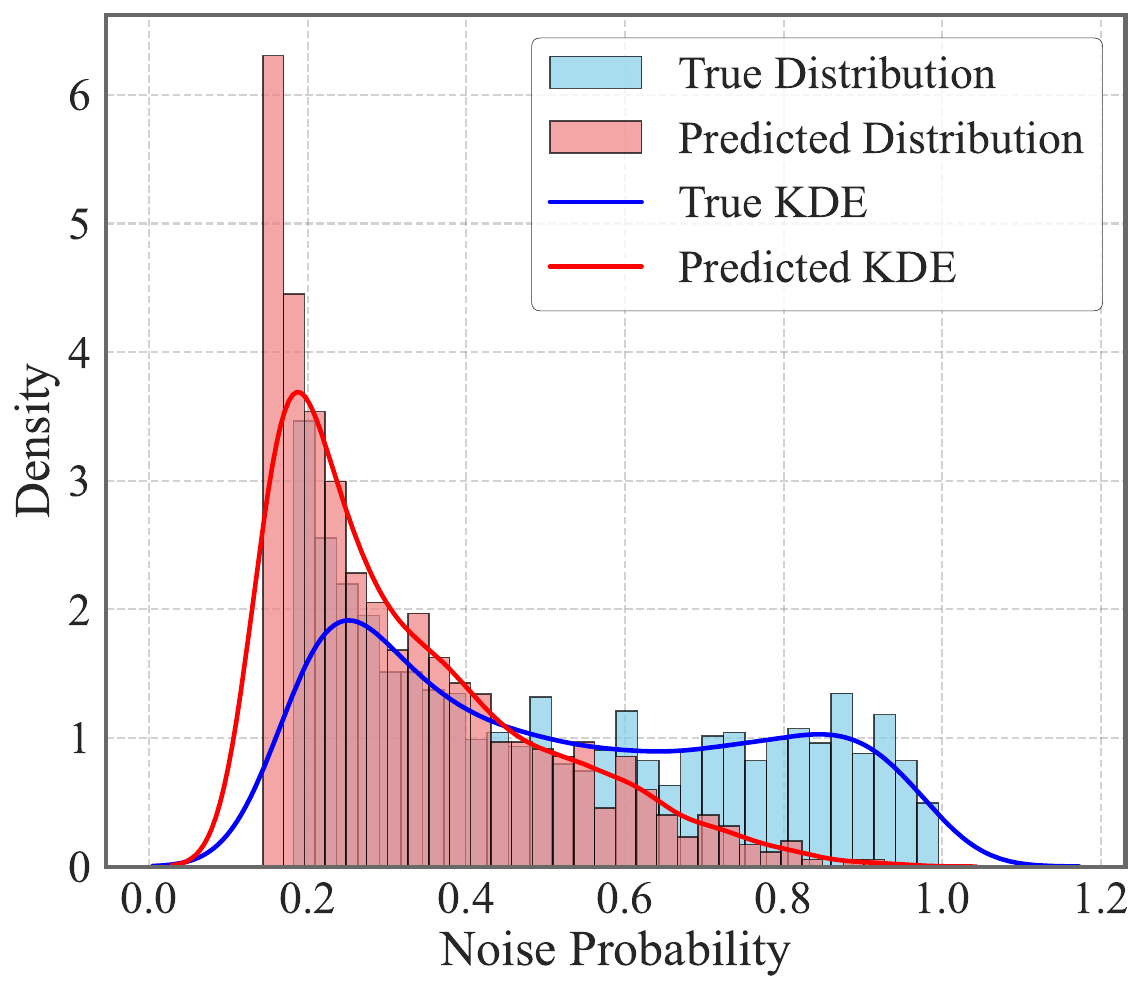}
        \end{subfigure}%
        \begin{subfigure}[b]{0.24\linewidth}
            \includegraphics[width=\textwidth]{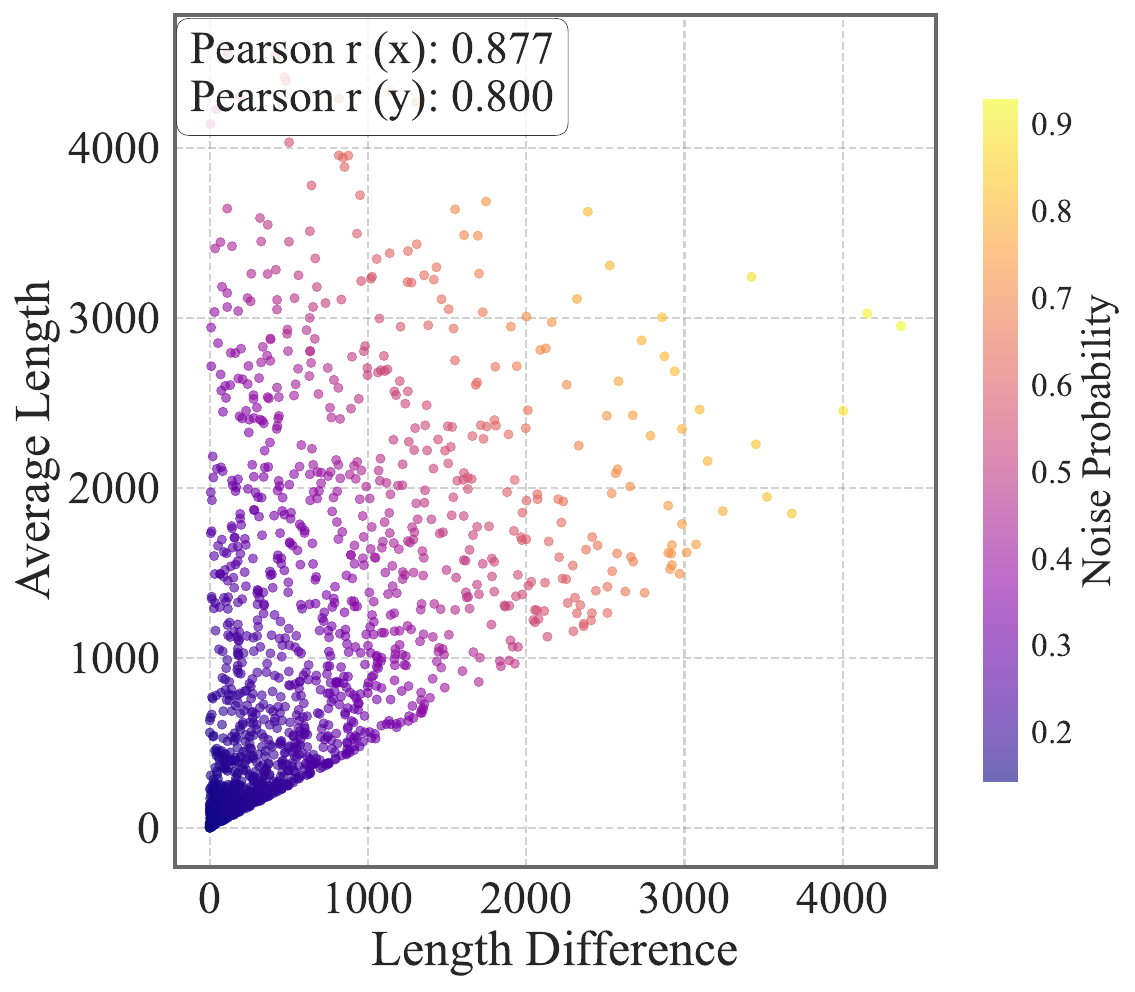}
        \end{subfigure}
        \caption*{{(b) Flipping Model Characterization at Flip Ratio 0.2}} 
    \end{minipage}
    \begin{minipage}{\linewidth}
        \centering
        \begin{subfigure}[b]{0.24\linewidth}
            \includegraphics[width=\textwidth]{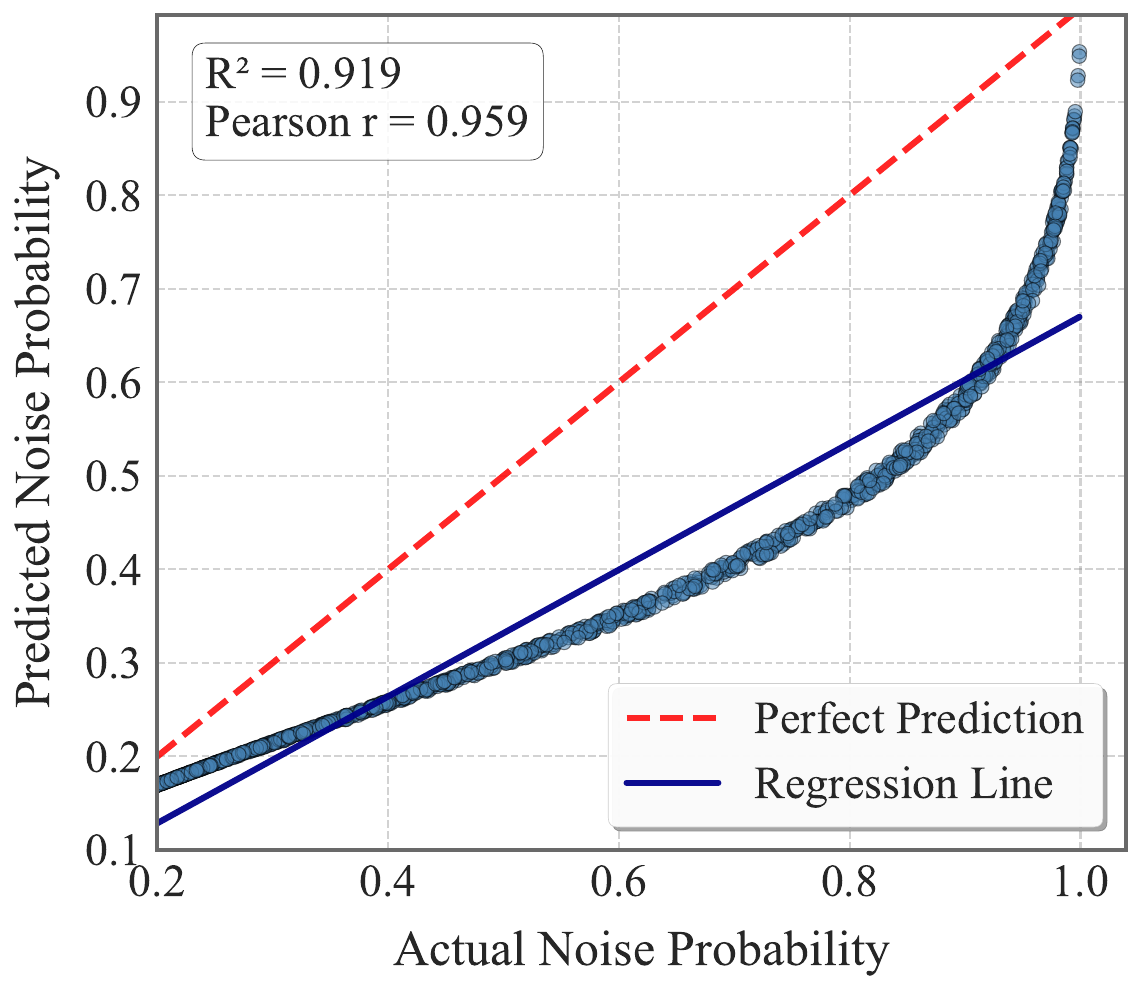}
        \end{subfigure}%
        \begin{subfigure}[b]{0.24\linewidth}
            \includegraphics[width=\textwidth]{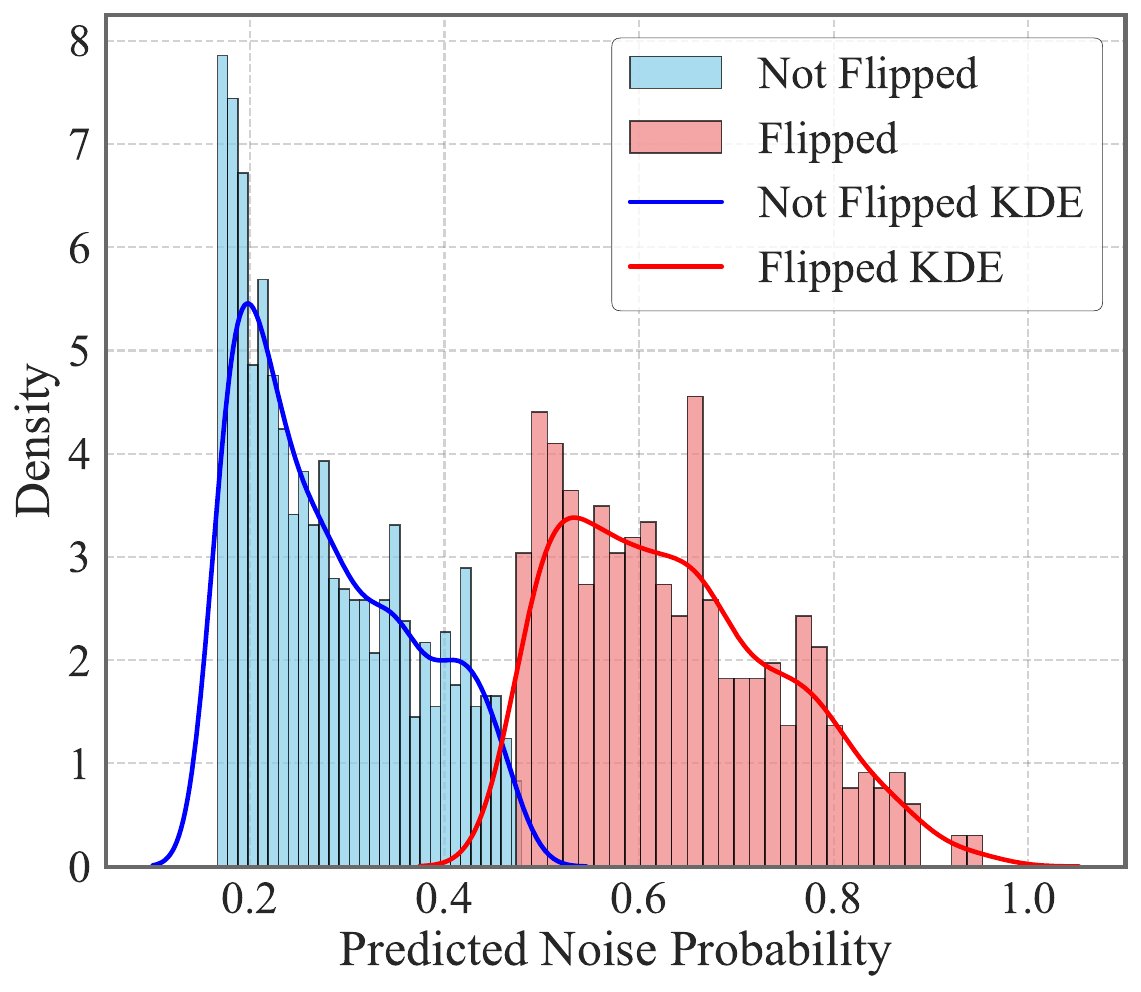}
        \end{subfigure}%
        \begin{subfigure}[b]{0.24\linewidth}
            \includegraphics[width=\textwidth]{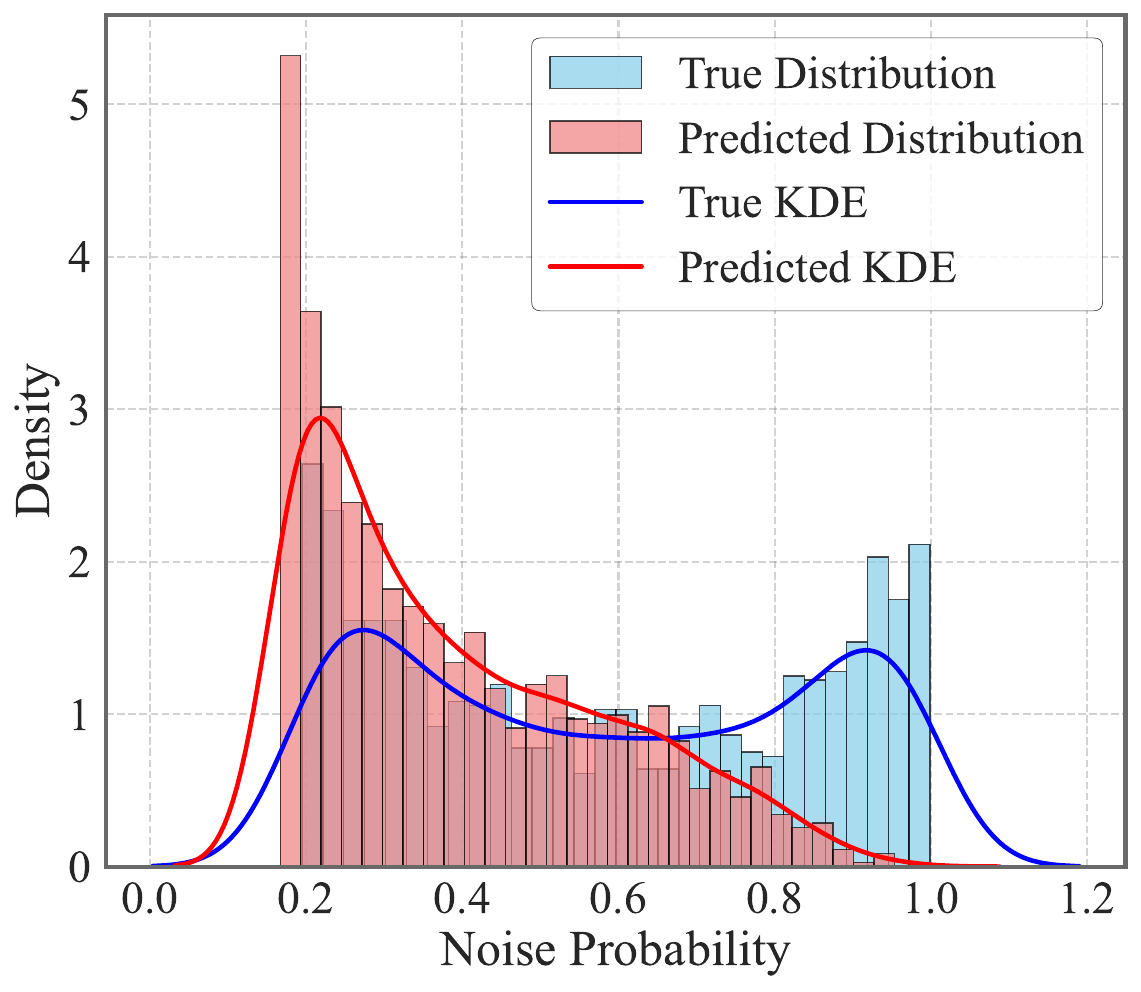}
        \end{subfigure}%
        \begin{subfigure}[b]{0.24\linewidth}
            \includegraphics[width=\textwidth]{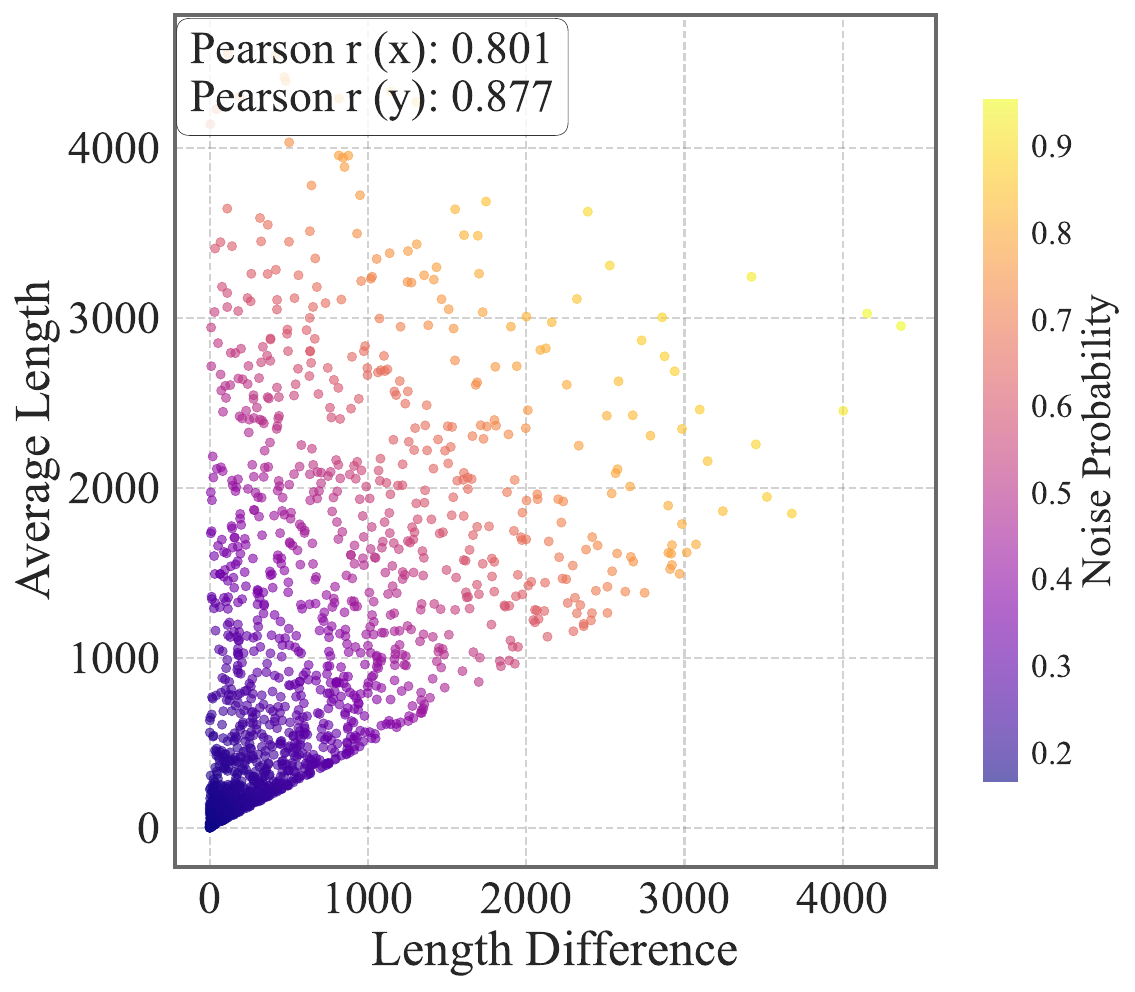}
        \end{subfigure}
        \caption*{{(c) Flipping Model Characterization at Flip Ratio 0.3}} 
    \end{minipage}
    \begin{minipage}{\linewidth}
        \centering
        \begin{subfigure}[b]{0.24\linewidth}
            \includegraphics[width=\textwidth]{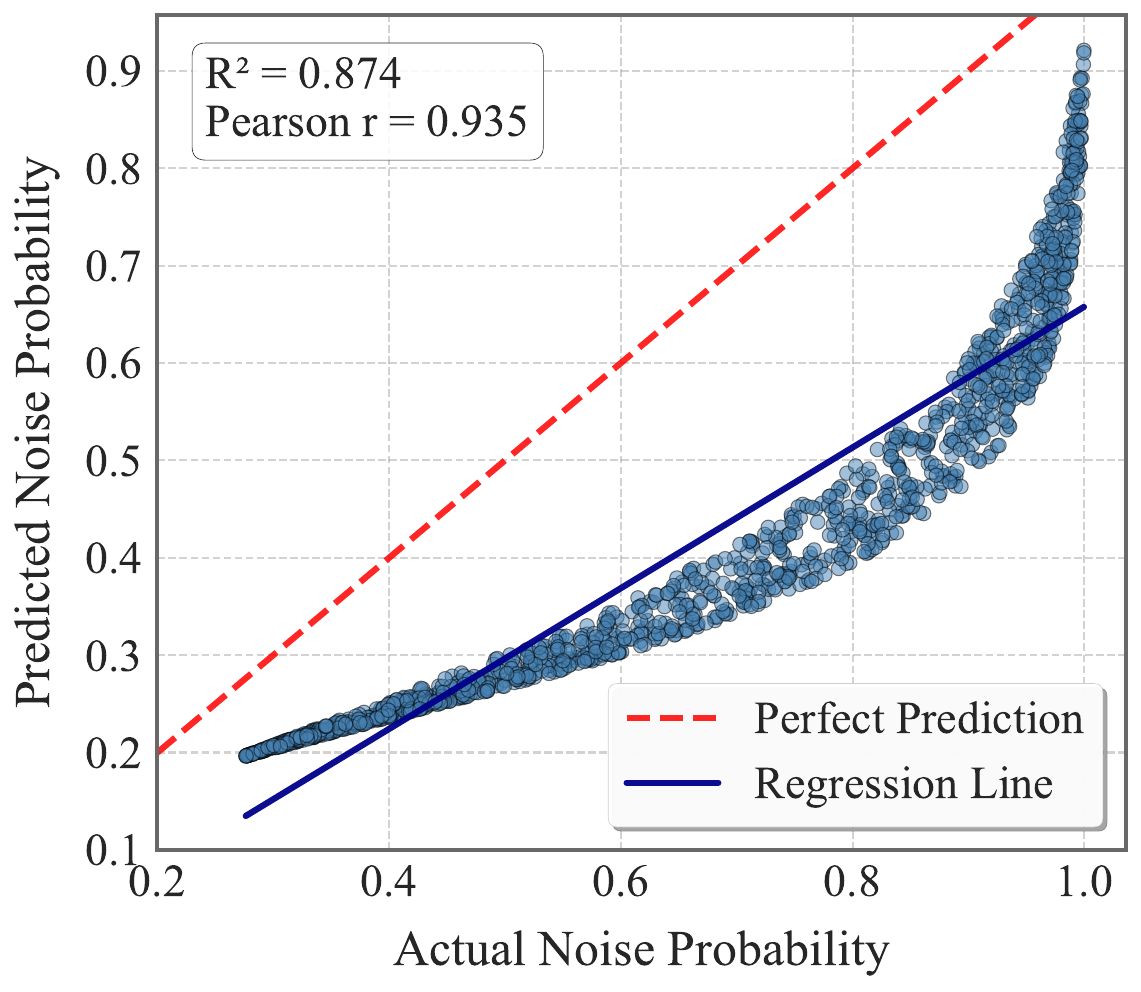}
        \end{subfigure}%
        \begin{subfigure}[b]{0.24\linewidth}
            \includegraphics[width=\textwidth]{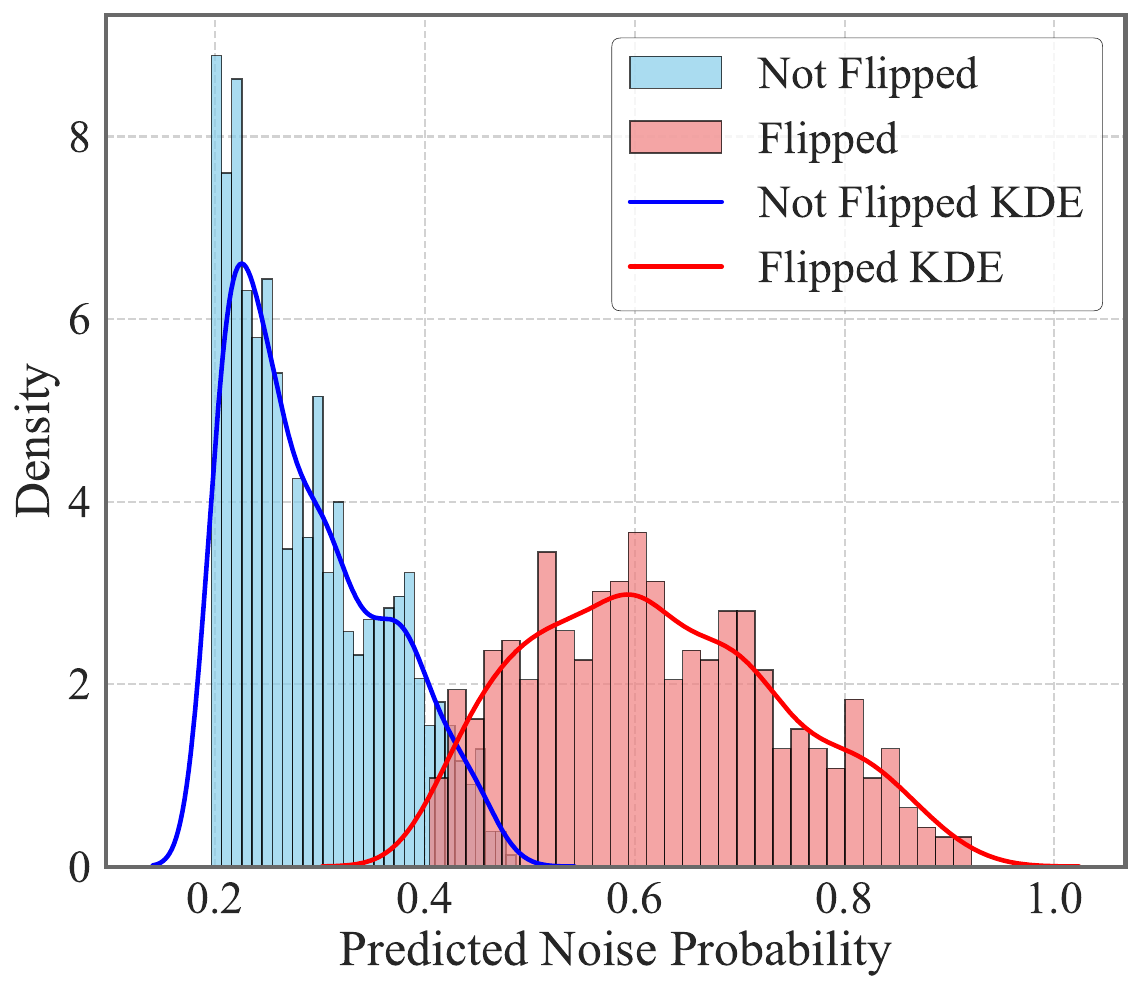}
        \end{subfigure}%
        \begin{subfigure}[b]{0.24\linewidth}
            \includegraphics[width=\textwidth]{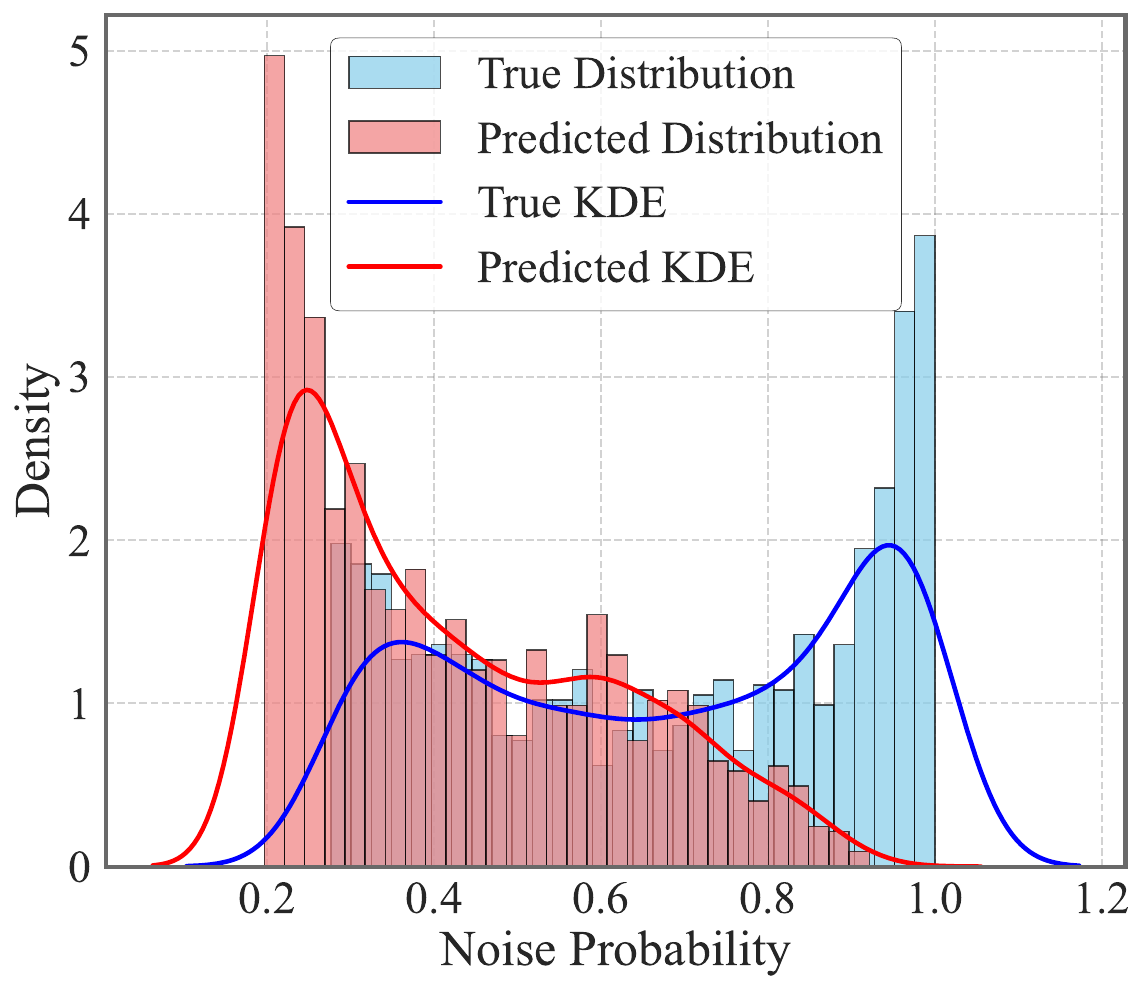}
        \end{subfigure}%
        \begin{subfigure}[b]{0.24\linewidth}
            \includegraphics[width=\textwidth]{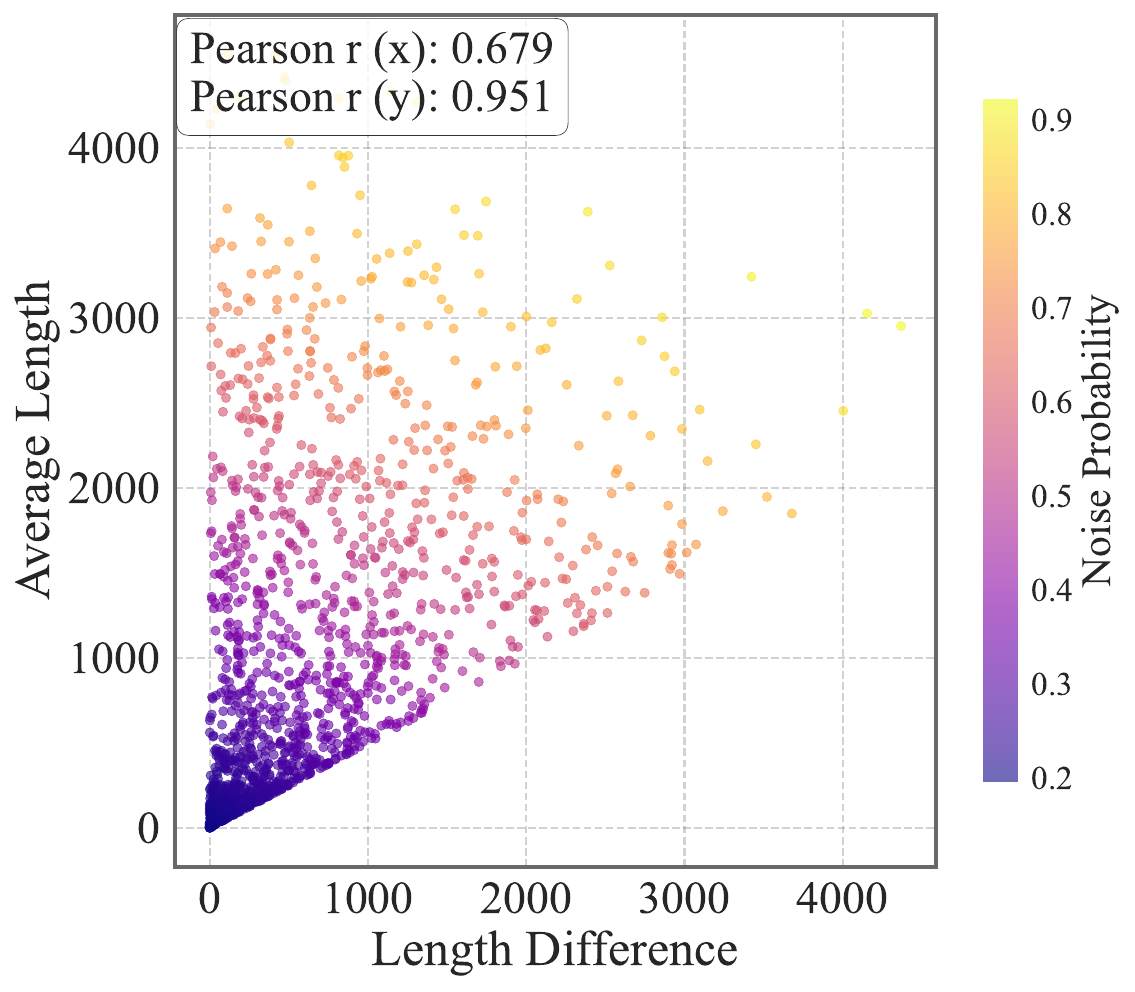}
        \end{subfigure}
        \caption*{{(d) Flipping Model Characterization at Flip Ratio 0.4}} 
    \end{minipage}

    \caption{Learned Flipping Model Characterization across 4 flip ratios} 
\end{figure}

\end{document}

%% file: math_commands.tex
\usepackage{amsmath,amsfonts,bm}

\def\eqref#1{equation~\ref{#1}}

\def\1{\bm{1}}

\def\vh{{\bm{h}}}

\def\vp{{\bm{p}}}

\def\vx{{\bm{x}}}
\def\vy{{\bm{y}}}

\DeclareMathAlphabet{\mathsfit}{\encodingdefault}{\sfdefault}{m}{sl}
\SetMathAlphabet{\mathsfit}{bold}{\encodingdefault}{\sfdefault}{bx}{n}

\def\sI{{\mathbb{I}}}

\def\sP{{\mathbb{P}}}

\newcommand{\E}{\mathbb{E}}

\newcommand{\sigmoid}{\sigma}

\DeclareMathOperator*{\argmin}{arg\,min}